\documentclass[conference]{IEEEtran}
\IEEEoverridecommandlockouts
\usepackage{cite}
\usepackage{amsmath,amssymb,amsfonts}
\usepackage{algorithmic}
\usepackage{graphicx}
\usepackage{textcomp}

\usepackage{multirow}
\usepackage{tikz} 
\usetikzlibrary{fit}
\usetikzlibrary{calc}
\usetikzlibrary{arrows,shapes,trees, patterns, backgrounds}
\usepackage{pgf}

\usepackage[utf8]{inputenc}

\ifCLASSOPTIONcompsoc 
	\usepackage[caption=false,font=normalsize,labelfont=sf,textfont=sf]{subfig}
\else
	\usepackage[caption=false,font=footnotesize]{subfig}

\def\BibTeX{{\rm B\kern-.05em{\sc i\kern-.025em b}\kern-.08em
    T\kern-.1667em\lower.7ex\hbox{E}\kern-.125emX}}

\begin{document}

\title{dhSegment: A generic deep-learning approach for document segmentation}

\author{
  \IEEEauthorblockN{
      \linespread{1.5}
	  Sofia Ares Oliveira\IEEEauthorrefmark{2}\IEEEauthorrefmark{1},
	  Benoit Seguin\IEEEauthorrefmark{2}\IEEEauthorrefmark{1},
	  Frederic Kaplan\IEEEauthorrefmark{2}
  }
  \vspace{1em}
  \IEEEauthorblockA{
	  \IEEEauthorrefmark{2}Digital Humanities Laboratory, EPFL,
	  Lausanne, VD, Switzerland \\
	  \textit {\{sofia.oliveiraares, benoit.seguin, frederic.kaplan\}@epfl.ch}
  }
  \vspace{1em}
  \IEEEauthorblockA{
      \IEEEauthorrefmark{1}These authors contributed equally to this work
  }
}

\maketitle

\begin{abstract}

In recent years there have been multiple successful attempts tackling document processing problems separately by designing task specific hand-tuned strategies. We argue that the diversity of historical document processing tasks prohibits to solve them one at a time and shows a need for designing generic approaches in order to handle the variability of historical series. In this paper, we address multiple tasks simultaneously such as page extraction, baseline extraction, layout analysis or multiple typologies of illustrations and photograph extraction. We propose an open-source implementation of a CNN-based pixel-wise predictor coupled with task dependent post-processing blocks. We show that a single CNN-architecture can be used across tasks with competitive results. Moreover most of the task-specific post-precessing steps can be decomposed in a small number of simple and standard reusable operations, adding to the flexibility of our approach. 

\end{abstract}


\section{Introduction}

When working with digitized historical documents, one is frequently faced with recurring needs and problems: how to cut out the page of the manuscript, how to extract the illustration from the text, how to find the pages that contain a certain type of symbol, how to locate text in a digitized image, etc. However, the domain of document analysis has been dominated for a long time by collections of heterogeneous segmentation methods, tailored for specific classes of problems and particular typologies of documents. We argue that the variability and diversity of historical series prevent us from tackling each problem separately, and that such specificity has been a great barrier towards off-the-shelf document analysis solutions, usable by non-specialists.

Lately, huge improvements have been made in semantic segmentation of natural images (roads, scenes, ...) but historical document processing and analysis have, in our opinion, not yet fully benefited from these. We believe that a tipping point has been reached and that recent progress in deep learning architectures may suggest that some generic approaches would be now mature enough to start outperforming dedicated systems. Also with the growing interest in digital humanities research, the need for simple-to-use, flexible and efficient tools to perform such analysis increases. 

This work is a contribution towards this goal and introduces dhSegment, a general and flexible architecture for pixel-wise segmentation related tasks on historical documents. We present the surprisingly good results of such a generic architecture across tasks common in historical document processing, and show that the proposed model is competitive or outperforming state-of-the-art methods. These encouraging results may have important consequences for the future of document analysis pipelines based on optimized generic building blocks. The implementation is open-source and available on Github\footnote{dhSegment implementation : https://github.com/dhlab-epfl/dhSegment}.

\section{Related Work}

In the recent years, Long et al. \cite{long2015fully} popularized the use of end-to-end fully convolutional networks (FCN) for semantic segmentation. They used an ImageNet \cite{deng2009imagenet} pretrained network, deconvolutional layers for upsampling and combined final prediction layer with lower layers (skip connections) to improve the predictions. The U-Net architecture \cite{ronneberger2015u} extended the FCN by setting the expansive path (decoder) to be symmetric to the contracting path (encoder), resulting in an u-shaped architecture with skip connections for each level.

Similarly, the architectures for Convolutional Neural Networks (CNN) have evolved drastically in the last years with architectures such as Alexnet \cite{krizhevsky2012imagenet}, VGG \cite{simonyan2014very} and ResNet \cite{he2016deep}. These architecture contributed greatly to the success and the massive usage of deep neural networks in many tasks and domains.


To some extent, historical document processing has also experienced the arrival of neural networks. As seen in the last competitions in document processing tasks \cite{simistira2017icdar2017, diem2017cbad, pratikakis2017icdar2017}, several successful methods make use of neural network approaches \cite{chen2017convolutional, xu2017page, breuel2017robust}, such as u-shaped and MDLSTM architectures for pixel-wise segmentation tasks.

\section{Approach}
\subsection{Outline}

The system is based on two successive steps which can be seen in Figure \ref{tikz:system}:

\begin{figure}
	\centering
	\input{content/system_schematics}
\end{figure}

\begin{itemize}
\item The first step is a Fully Convolutional Neural Network which takes as input the image of the document to be processed and outputs a map of probabilities of attributes predicted for each pixel. 
Training labels are used to generate masks and these mask images constitute the input data to train the network.
\item The second step transforms the map of predictions to the desired output of the task. We only allow ourselves simple standard image processing techniques, which are task dependent because of the diversity of outputs required.
\end{itemize}

The implementation of the network uses TensorFlow \cite{tensorflow2015-whitepaper}. 

\subsection{Network architecture}

The architecture of the network is depicted in Figure \ref{tikz:network}. dhSegment is composed of a contracting path\footnote{We reuse the terminology 'contracting' and 'exapanding' paths of \cite{ronneberger2015u}}, which follows the deep residual network ResNet-50 \cite{he2016deep} architecture (yellow blocks), and a expansive path that maps the low resolution encoder feature maps to full input resolution feature maps. 
Each path has five steps corresponding to five feature maps' sizes $S$, each step $i$ halving the previous step's feature maps size. 




\pgfmathsetmacro{\xsep}{0.2}
\pgfmathsetmacro{\ysep}{0.3}
\pgfmathsetmacro{\hI}{1.9} 
\pgfmathsetmacro{\wI}{0.02} 
\pgfmathsetmacro{\wId}{0.06} 
\pgfmathsetmacro{\hII}{1.2} 
\pgfmathsetmacro{\wII}{0.09} 
\pgfmathsetmacro{\hIII}{0.7} 
\pgfmathsetmacro{\wIId}{0.14} 
\pgfmathsetmacro{\wIII}{0.18} 
\pgfmathsetmacro{\hIV}{0.4} 
\pgfmathsetmacro{\wIV}{0.32} 
\pgfmathsetmacro{\hV}{0.25} 
\pgfmathsetmacro{\wV}{0.58} 
\pgfmathsetmacro{\hVI}{0.15} 
\pgfmathsetmacro{\wVI}{1} 
\pgfmathsetmacro{\shiftlabel}{0.25} 
\pgfmathsetmacro{\sizearrow}{0.3} 
\pgfmathsetmacro{\hseplegend}{0.23} 

\definecolor{mygrayfill}{RGB}{200,200,200}
\definecolor{mygraydraw}{RGB}{180,180,180}
\definecolor{mybluedashedfill}{RGB}{102,178,255}
\definecolor{mybluedasheddraw}{RGB}{51,153,255}
\definecolor{myyellowfill}{RGB}{255,255,204}
\definecolor{myyellowdraw}{RGB}{255,255,51}
\definecolor{myorangetext}{RGB}{255,153,51}

\definecolor{myvioletarrow}{RGB}{178,102,255}
\definecolor{mygreenarrow}{RGB}{0,205,0}
\definecolor{myredarrow}{RGB}{255,51,51}
\definecolor{mycyanarrow}{RGB}{0,204,204}
\definecolor{myorangearrow}{RGB}{255,128,0}
\definecolor{mypinkarrow}{RGB}{255,204,229}
\definecolor{mybluearrow}{RGB}{0,102,204}
\definecolor{mydashedarrow}{RGB}{255,178,102}

\begin{figure*}
\centering
\begin{tikzpicture}[
myrect/.style={
  rectangle,
  draw=mygraydraw,
  fill=mygrayfill,
  inner sep=0pt,
  fit=#1},
mydashedrect/.style={
  rectangle,
  draw=mybluedasheddraw,
  pattern=north west lines, 
  pattern color=mybluedashedfill, 
  dashed,
  inner sep=0pt, 
  fit=#1},
myinvisiblerect/.style={
  rectangle,
  draw=none,
  inner sep=0pt, 
  fit=#1},
myblockrect/.style={
  rectangle,
  draw=myyellowdraw,
  fill=myyellowfill,
  dashed,
  inner sep=0pt, 
  fit=#1},
upsampling/.style={
	->,
	mygreenarrow,
	thick},
bottleneck/.style={
	->,
	myvioletarrow,
	thick},
bottleneck_s2/.style={
	->,
	myredarrow,
	thick},
conv1x1/.style={
	->,
	mycyanarrow,
	thick},
conv3x3/.style={
	->,
	myorangearrow,
	thick},
conv7x7_s2/.style={
	->,
	mypinkarrow,
	thick},
maxpool2/.style={
	->,
	mybluearrow,
	thick},
copy/.style={
	->,
	mydashedarrow, 
	dashed},
labelfilters/.style 2 args={
	label={[black!60, shift={(#1,0.1)}]above:{\tiny #2}}}
][y=-1cm]

	\begin{scope}[y=-1cm]
		\coordinate (A_origin) at (0, 0);
		\coordinate (B_input) at ([shift={(\wI, \hI)}]A_origin);
		\node[myrect={(A_origin) (B_input)}, labelfilters={0}{3}] (input) {};
		
		\coordinate (A_conv) at ([shift={(0, \hI + \ysep)}]A_origin);
		\coordinate (B_conv) at ([shift={(\wII, \hII)}]A_conv);
		\node[myrect={(A_conv) (B_conv)}, labelfilters={-\shiftlabel}{64}] (conv1) {};
		
		\coordinate (A_pool) at ([shift={(0, \hII + \ysep)}]A_conv); 
		\coordinate (B_pool) at ([shift={(\wII, \hIII)}]A_pool); 
		\node[myrect={(A_pool) (B_pool)}, labelfilters={-\shiftlabel}{64}] (pool1) {};
		
		\coordinate (A_b1u1) at ([shift={(\wII + \xsep, 0)}]A_pool); 
		\coordinate (B_b1u1) at ([shift={(\wIII, \hIII)}]A_b1u1); 
		\node[myrect={(A_b1u1) (B_b1u1)}, labelfilters={0}{256}] (b1_u1) {};
		
		\coordinate (A_b1u2) at ([shift={(\wIII + \xsep, 0)}]A_b1u1); 
		\coordinate (B_b1u2) at ([shift={(\wIII, \hIII)}]A_b1u2); 
		\node[myrect={(A_b1u2) (B_b1u2)}, labelfilters={0}{256}] (b1_u2) {};
		
		\coordinate (A_b1u3) at ([shift={(0, \hIII + \ysep)}]A_b1u2); 
		\coordinate (B_b1u3) at ([shift={(\wIII, \hIV)}]A_b1u3); 
		\node[myrect={(A_b1u3) (B_b1u3)}, labelfilters={-\shiftlabel}{256}] (b1_u3) {};
		
		\coordinate (A_b2u1) at ([shift={(\wIII + \xsep, 0)}]A_b1u3); 
		\coordinate (B_b2u1) at ([shift={(\wIV, \hIV)}]A_b2u1); 
		\node[myrect={(A_b2u1) (B_b2u1)}, labelfilters={0}{512}] (b2_u1) {};
		
		\coordinate (A_b2u2) at ([shift={(\wIII + 2*\xsep, 0)}]A_b2u1); 
		\coordinate (B_b2u2) at ([shift={(\wIV, \hIV)}]A_b2u2); 
		\node[myrect={(A_b2u2) (B_b2u2)}, labelfilters={0}{512}] (b2_u2) {};
		
		\coordinate (A_b2u3) at ([shift={(\wIII + 2*\xsep, 0)}]A_b2u2); 
		\coordinate (B_b2u3) at ([shift={(\wIV, \hIV)}]A_b2u3); 
		\node[myrect={(A_b2u3) (B_b2u3)}, labelfilters={0}{512}] (b2_u3) {};
		
		\coordinate (A_b2u4) at ([shift={(0, \hIV + \ysep)}]A_b2u3); 
		\coordinate (B_b2u4) at ([shift={(\wIV, \hV)}]A_b2u4); 
		\node[myrect={(A_b2u4) (B_b2u4)}, labelfilters={-\shiftlabel}{512}] (b2_u4) {};
		
		\coordinate (A_b3u1) at ([shift={(\wIV + \xsep, 0)}]A_b2u4); 
		\coordinate (B_b3u1) at ([shift={(\wV, \hV)}]A_b3u1); 
		\node[myrect={(A_b3u1) (B_b3u1)}, labelfilters={0}{1024}] (b3_u1) {};
		
		\coordinate (A_b3u2) at ([shift={(\wV + \xsep, 0)}]A_b3u1); 
		\coordinate (B_b3u2) at ([shift={(\wV, \hV)}]A_b3u2); 
		\node[myrect={(A_b3u2) (B_b3u2)}, labelfilters={0}{1024}] (b3_u2) {};
		
		\coordinate (A_b3u3) at ([shift={(\wV + \xsep, 0)}]A_b3u2); 
		\coordinate (B_b3u3) at ([shift={(\wV, \hV)}]A_b3u3); 
		\node[myrect={(A_b3u3) (B_b3u3)}, labelfilters={0}{1024}] (b3_u3) {};
		
		\coordinate (A_b3u4) at ([shift={(\wV + \xsep, 0)}]A_b3u3); 
		\coordinate (B_b3u4) at ([shift={(\wV, \hV)}]A_b3u4); 
		\node[myrect={(A_b3u4) (B_b3u4)}, labelfilters={0}{1024}] (b3_u4) {};
		
		\coordinate (A_b3u5) at ([shift={(\wV + \xsep, 0)}]A_b3u4); 
		\coordinate (B_b3u5) at ([shift={(\wV, \hV)}]A_b3u5); 
		\node[myrect={(A_b3u5) (B_b3u5)}, labelfilters={0}{1024}] (b3_u5) {};
		
		\coordinate (A_b3u6) at ([shift={(0, \hV + \ysep)}]A_b3u5); 
		\coordinate (B_b3u6) at ([shift={(\wV, \hVI)}]A_b3u6); 
		\node[myrect={(A_b3u6) (B_b3u6)}, labelfilters={-\shiftlabel-0.05}{1024}] (b3_u6) {};
				
		\coordinate (A_b3dimred) at ([shift={(\wV + 1.5*\xsep, 0)}]A_b3u5); 
		\coordinate (B_b3dimred) at ([shift={(\wV/2, \hV)}]A_b3dimred); 
		\node[myrect={(A_b3dimred) (B_b3dimred)}, labelfilters={0}{512}] (b3_dimred) {};
		
		\coordinate (A_b4u1) at ([shift={(\wV + \xsep, 0)}]A_b3u6); 
		\coordinate (B_b4u1) at ([shift={(\wVI, \hVI)}]A_b4u1); 
		\node[myrect={(A_b4u1) (B_b4u1)}, labelfilters={0}{2048}] (b4_u1) {};
		
		\coordinate (A_b4u2) at ([shift={(\wVI + \xsep, 0)}]A_b4u1); 
		\coordinate (B_b4u2) at ([shift={(\wVI, \hVI)}]A_b4u2); 
		\node[myrect={(A_b4u2) (B_b4u2)}, labelfilters={0}{2048}] (b4_u2) {};
		
		\coordinate (A_b4u3) at ([shift={(\wVI + \xsep, 0)}]A_b4u2); 
		\coordinate (B_b4u3) at ([shift={(\wVI, \hVI)}]A_b4u3); 
		\node[myrect={(A_b4u3) (B_b4u3)}, labelfilters={0}{2048}] (b4_u3) {};
			
		\coordinate (A_b4dimred) at ([shift={(\wVI + 1.5*\xsep, 0)}]A_b4u3); 
		\coordinate (B_b4dimred) at ([shift={(\wIV, \hVI)}]A_b4dimred); 
		\node[myrect={(A_b4dimred) (B_b4dimred)}, labelfilters={\shiftlabel}{512}] (b4_dimred) {};
		
		\coordinate (A_up1) at ([shift={(0, -(\ysep + \hV))}]A_b4dimred); 
		\coordinate (B_up1) at ([shift={(\wIV, \hV)}]A_up1); 
		\node[myrect={(A_up1) (B_up1)}, labelfilters={0}{512}] (up1) {};
		\coordinate (A_cat1) at ([shift={(-\wIV, 0)}]A_up1); 
		\coordinate (B_cat1) at ([shift={(\wIV, \hV)}]A_cat1); 
		\node[mydashedrect={(A_cat1) (B_cat1)}] (cat1) {};
		\coordinate (A_deconv1) at ([shift={(\wIV + \xsep, 0)}]A_up1); 
		\coordinate (B_deconv1) at ([shift={(\wIV, \hV)}]A_deconv1); 
		\node[myrect={(A_deconv1) (B_deconv1)}, labelfilters={\shiftlabel}{512}] (deconv1) {};
		
		\coordinate (A_up2) at ([shift={(0, -(\ysep + \hIV))}]A_deconv1); 
		\coordinate (B_up2) at ([shift={(\wIV, \hIV)}]A_up2); 
		\node[myrect={(A_up2) (B_up2)}, labelfilters={0}{512}] (up2) {};
		\coordinate (A_cat2) at ([shift={(-\wIV, 0)}]A_up2); 
		\coordinate (B_cat2) at ([shift={(\wIV, \hIV)}]A_cat2); 
		\node[mydashedrect={(A_cat2) (B_cat2)}] (cat2) {};
		\coordinate (A_deconv2) at ([shift={(\wIV + \xsep, 0)}]A_up2); 
		\coordinate (B_deconv2) at ([shift={(\wIII, \hIV)}]A_deconv2); 
		\node[myrect={(A_deconv2) (B_deconv2)}, labelfilters={\shiftlabel}{256}] (deconv2) {};
		
		\coordinate (A_up3) at ([shift={(0, -(\ysep + \hIII))}]A_deconv2); 
		\coordinate (B_up3) at ([shift={(\wIII, \hIII)}]A_up3); 
		\node[myrect={(A_up3) (B_up3)}, labelfilters={0}{256}] (up3) {};
		\coordinate (A_cat3) at ([shift={(-\wIII, 0)}]A_up3); 
		\coordinate (B_cat3) at ([shift={(\wIII, \hIII)}]A_cat3); 
		\node[mydashedrect={(A_cat3) (B_cat3)}] (cat3) {};
		\coordinate (A_deconv3) at ([shift={(\wIII + \xsep, 0)}]A_up3); 
		\coordinate (B_deconv3) at ([shift={(\wIId, \hIII)}]A_deconv3); 
		\node[myrect={(A_deconv3) (B_deconv3)}, labelfilters={\shiftlabel}{128}] (deconv3) {};
		
		\coordinate (A_up4) at ([shift={(0, -(\ysep + \hII))}]A_deconv3); 
		\coordinate (B_up4) at ([shift={(\wIId, \hII)}]A_up4); 
		\node[myrect={(A_up4) (B_up4)}, labelfilters={0}{128}] (up4) {};
		\coordinate (A_cat4) at ([shift={(-\wII, 0)}]A_up4); 
		\coordinate (B_cat4) at ([shift={(\wII, \hII)}]A_cat4); 
		\node[mydashedrect={(A_cat4) (B_cat4)}] (cat4) {};
		\coordinate (A_deconv4) at ([shift={(\wIId + \xsep, 0)}]A_up4); 
		\coordinate (B_deconv4) at ([shift={(\wII, \hII)}]A_deconv4); 
		\node[myrect={(A_deconv4) (B_deconv4)}, labelfilters={\shiftlabel}{64}] (deconv4) {};
		
		\coordinate (A_up5) at ([shift={(0, -(\ysep + \hI))}]A_deconv4); 
		\coordinate (B_up5) at ([shift={(\wII, \hI)}]A_up5); 
		\node[myrect={(A_up5) (B_up5)}, labelfilters={0}{64}] (up5) {};
		\coordinate (A_cat5) at ([shift={(-\wI, 0)}]A_up5); 
		\coordinate (B_cat5) at ([shift={(\wI, \hI)}]A_cat5); 
		\node[mydashedrect={(A_cat5) (B_cat5)}] (cat5) {};
		\coordinate (A_deconv5) at ([shift={(\wId + \xsep, 0)}]A_up5); 
		\coordinate (B_deconv5) at ([shift={(\wId, \hI)}]A_deconv5); 
		\node[myrect={(A_deconv5) (B_deconv5)}, labelfilters={0}{32}] (deconv5) {};
		
		\coordinate (A_out) at ([shift={(\wId + 1.5*\xsep, 0)}]A_deconv5); 
		\coordinate (B_out) at ([shift={(\wI, \hI)}]A_out); 
		\node[myrect={(A_out) (B_out)}, labelfilters={0}{c}] (output) {};
			
		\draw[conv7x7_s2] (input) -- (conv1);
		\draw[maxpool2] (conv1) -- (pool1);
		\draw[bottleneck] (pool1) -- (b1_u1);
		
		\draw[bottleneck] (b1_u1) -- (b1_u2);
		\draw[bottleneck_s2] (b1_u2) -- (b1_u3);
		\draw[bottleneck] (b1_u3) -- (b2_u1);
		
		\draw[bottleneck] (b2_u1) -- (b2_u2);
		\draw[bottleneck] (b2_u2) -- (b2_u3);
		\draw[bottleneck_s2] (b2_u3) -- (b2_u4);
		\draw[bottleneck] (b2_u4) -- (b3_u1);
		
		\draw[bottleneck] (b3_u1) -- (b3_u2);
		\draw[bottleneck] (b3_u2) -- (b3_u3);
		\draw[bottleneck] (b3_u3) -- (b3_u4);
		\draw[bottleneck] (b3_u4) -- (b3_u5);
		\draw[bottleneck_s2] (b3_u5) -- (b3_u6);
		\draw[bottleneck] (b3_u6) -- (b4_u1);
		
		\draw[bottleneck] (b4_u1) -- (b4_u2);
		\draw[bottleneck] (b4_u2) -- (b4_u3);
		
		\draw[upsampling] (b4_dimred) -- (up1);
		\draw[conv3x3] (up1) -- (deconv1);
		
		\draw[upsampling] (deconv1) -- (up2);
		\draw[conv3x3] (up2) -- (deconv2);
		
		\draw[upsampling] (deconv2) -- (up3);
		\draw[conv3x3] (up3) -- (deconv3);
		
		\draw[upsampling] (deconv3) -- (up4);
		\draw[conv3x3] (up4) -- (deconv4);
		
		\draw[upsampling] (deconv4) -- (up5);
		\draw[conv3x3] (up5) -- (deconv5);
		
		\draw[copy] (b3_dimred) -- (cat1);
		\draw[copy] (b2_u3) -- (cat2);
		\draw[copy] (b1_u2) -- (cat3);
		\draw[copy] (conv1) -- (cat4);
		\draw[copy] (input) -- (cat5);
		
		\draw[conv1x1] (b4_u3) -- (b4_dimred);
		\draw[conv1x1] (b3_u5) -- (b3_dimred);
		\draw[conv1x1] (deconv5) -- (output);

		\coordinate (A_s) at ([shift={(0.5*\xsep, 0)}]A_out); 
		\coordinate (B_s) at ([shift={(0, \hI)}]A_s); 
		\node [myinvisiblerect={(A_s) (B_s)}, label={[gray]right:{\tiny S}}] {};
		
		\coordinate (A_s2) at ([shift={(0, \hI + \ysep)}]A_s); 
		\coordinate (B_s2) at ([shift={(0, \hII)}]A_s2); 
		\node [myinvisiblerect={(A_s2) (B_s2)}, label={[gray]right:{\tiny S/2}}] {};
		
		\coordinate (A_s4) at ([shift={(0, \hII + \ysep)}]A_s2); 
		\coordinate (B_s4) at ([shift={(0, \hIII)}]A_s4); 
		\node [myinvisiblerect={(A_s4) (B_s4)}, label={[gray]right:{\tiny S/4}}] {};
		
		\coordinate (A_s8) at ([shift={(0, \hIII + \ysep)}]A_s4); 
		\coordinate (B_s8) at ([shift={(0, \hIV)}]A_s8); 
		\node [myinvisiblerect={(A_s8) (B_s8)}, label={[gray]right:{\tiny S/8}}] {};
		
		\coordinate (A_s16) at ([shift={(0, \hIV + \ysep)}]A_s8); 
		\coordinate (B_s16) at ([shift={(0, \hV)}]A_s16); 
		\node [myinvisiblerect={(A_s16) (B_s16)}, label={[gray]right:{\tiny S/16}}] {};
		
		\coordinate (A_s32) at ([shift={(0, \hV + \ysep)}]A_s16); 
		\coordinate (B_s32) at ([shift={(0, \hVI)}]A_s32); 
		\node [myinvisiblerect={(A_s32) (B_s32)}, label={[gray]right:{\tiny S/32}}] {};
		
		\begin{scope}[on background layer]
			\coordinate (A_B0) at ([shift={(-0.07, -0.1)}]A_conv); 
			\coordinate (B_B0) at ([shift={(0.07, 0.1)}]B_pool); 		
		
			\coordinate (A_B1) at ([shift={(-0.07, -0.2)}]A_b1u1); 
			\coordinate (B_B1) at ([shift={(0.07, 0.2)}]B_b1u3); 
			
			\coordinate (A_B2) at ([shift={(-0.05, -0.2)}]A_b2u1); 
			\coordinate (B_B2) at ([shift={(0.07, 0.1)}]B_b2u4); 
			
			\coordinate (A_B3) at ([shift={(-0.05, -0.2)}]A_b3u1); 
			\coordinate (B_B3) at ([shift={(0.07, 0.1)}]B_b3u6); 
			
			\coordinate (A_B4) at ([shift={(-0.05, -0.2)}]A_b4u1); 
			\coordinate (B_B4) at ([shift={(0.07, 0.25)}]B_b4u3);

			\node [myblockrect={(A_B0) (B_B0)}, label={[myorangetext, shift={(-0.2,-0.1)}]south:{\tiny Block0}}] {};
			\node [myblockrect={(A_B1) (B_B1)}, label={[myorangetext, shift={(0,-0.3)}]south:{\tiny Block1}}] {};
			\node [myblockrect={(A_B2) (B_B2)}, label={[myorangetext, shift={(-0.4,-0.3)}]south:{\tiny Block2}}] {};
			\node [myblockrect={(A_B3) (B_B3)}, label={[myorangetext, shift={(-1.5,-0.3)}]south:{\tiny Block3}}] {};
			\node [myblockrect={(A_B4) (B_B4)}, label={[myorangetext, shift={(-1.3,-0.3)}]south:{\tiny Block4}}] {};
			
		\end{scope}
		
	\end{scope}

\coordinate (O_legend) at (13.7,-3);
\draw [conv7x7_s2] (O_legend) to ([shift={(+\sizearrow, 0)}]O_legend) node[right] {\tiny conv 7x7 s/2};
\draw [maxpool2] ([shift={(0, -\hseplegend)}]O_legend) to ([shift={(+\sizearrow, -\hseplegend)}]O_legend) node[right] {\tiny max pooling 2x2};
\draw [bottleneck] ([shift={(0, -2*\hseplegend)}]O_legend) to ([shift={(+\sizearrow, -2*\hseplegend)}]O_legend) node[right] {\tiny bottleneck};
\draw [bottleneck_s2] ([shift={(0, -3*\hseplegend)}]O_legend) to ([shift={(+\sizearrow, -3*\hseplegend)}]O_legend) node[right] {\tiny bottleneck s/2};
\draw [conv1x1] ([shift={(0, -4*\hseplegend)}]O_legend) to ([shift={(+\sizearrow, -4*\hseplegend)}]O_legend) node[right] {\tiny conv1x1};
\draw [upsampling] ([shift={(0, -5*\hseplegend)}]O_legend) to ([shift={(+\sizearrow, -5*\hseplegend)}]O_legend) node[right] {\tiny upscaling};
\draw [conv3x3] ([shift={(0, -6*\hseplegend)}]O_legend) to ([shift={(+\sizearrow, -6*\hseplegend)}]O_legend) node[right] {\tiny conv3x3};
\draw [copy] ([shift={(0, -7*\hseplegend)}]O_legend) to ([shift={(+\sizearrow, -7*\hseplegend)}]O_legend) node[right] {\tiny copy};

\end{tikzpicture}
\caption{Network architecture of dhSegment. The yellow blocks correspond to ResNet-50 architecture which implementation is slightly different from the original in \cite{he2016deep} for memory efficiency reasons. The number of features channels are restricted to $512$ in the expansive path in order to limit the number of training parameters, thus the dimensionality reduction in the contracting path (light blue arrows). The dashed rectangles correspond to the copies of features maps from the contracting path that are concatenated with the up-sampled features maps of the expanding path. Each expanding step doubles the feature map's size and halves the number of features channels. The output prediction has the same size as the input image and the number of features channels constitute the desired number of classes.
} 
\label{tikz:network}
\end{figure*}
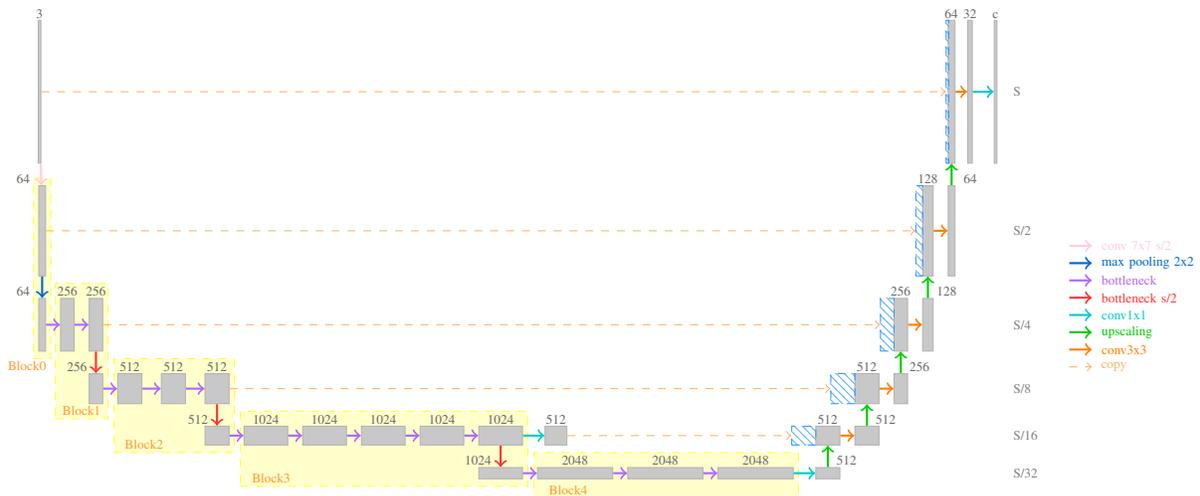


The contracting path uses pretrained weights as it adds robustness and helps generalization. It takes advantage of the high level features learned on a general image classification task (ImageNet \cite{deng2009imagenet}). For simplicity reasons the so-called ``bottleneck" blocks are shown as violet arrows and downsampling bottlenecks as red arrows in Figure \ref{tikz:network}. We refer the reader to \cite{he2016deep} for a detailed presentation of ResNet architecture.

The expanding path is composed of five blocks plus a final convolutional layer which assigns a class to each pixel. Each deconvolutional step is composed of an upscaling of the previous block feature map, a concatenation of the upscaled feature map with a copy of the corresponding contracting feature map and a 3x3 convolutional layer followed by a rectified linear unit (ReLU) \cite{nair2010rectified}. The number of features channels in step $i=4$ and $i=5$ are reduced to 512 by a 1x1 convolution before concatenation in order to reduce the number of parameters and memory usage. The upsampling is performed using a bilinear interpolation.

The architecture contains 32.8M parameters in total but since most of them are part of the pre-trained encoder, only 9.36M have to be fully-trained.\footnote{Actually one could argue that the 1.57M parameters coming from the dimensionality reduction blocks do not have to be fully trained either, thus reducing the number of fully-trainable parameters to 7.79M. Indeed, they are initialized as random projections, which is a valid way of reducing dimensionality, 
so they can also be considered as part of the fine-tuning of the pre-trained network.}

\subsection{Post-processing}
Our general approach to demonstrate the effectiveness and genericity of our network is to limit the post-processing steps to simple and standards operations on the predictions. 

\subsubsection*{Thresholding}
Thresholding is used to obtain a binary map from the predictions output by the network. If several classes are to be found, the thresholding is done class-wise. The threshold is either a fixed constant ($t \in [0,1]$) or found by Otsu's method \cite{otsu1979threshold}.

\subsubsection*{Morphological operations} 
Morphological operations are non-linear operations that originate from mathematical morphology theory \cite{serra1983image}. They are standard and widely used methods in image processing to analyse and process geometrical structures. The two fundamental basic operators, namely the erosion and dilation, can be combined to result in opening and closing operators. We limit our post-processing to these two operators applied on binary images.  

\subsubsection*{Connected components analysis}
In our case, connected components analysis is used in order to filter out small connected components that may remain after thresholding or morphological operations.

\subsubsection*{Shape vectorization} 
A vectorisation step is needed in order to transform the detected region into a set of coordinates. To do so, the blobs in the binary image are extracted as polygonal shapes. In fact, the polygons are usually bounding boxes represented by four corner points, which may be the minimum rectangle enclosing the object or quadrilaterals. The detected shape can also be a line and in this case, the vectorization consists in a path reduction.

\subsection{Training}
The training is regularized using L2 regularization with weight decay ($10^{-6}$). We use a learning rate with an exponential decay rate of $0.95$ and an initial value in $[10^{-5}, 10^{-4}]$. Xavier initialization \cite{glorot2010understanding} and Adam optimizer \cite{kingma2014adam} are used. 
Batch renormalization \cite{ioffe2017batch} is used to ensure that the lack of diversity in a given batch is not an issue (with $r_{min}=0.1$, $r_{max}=100$, $d_{max}=1$ ). 
The images are resized so that the total number of pixels lies between $6\cdot 10^5$ and $10^6$. Images are also cropped into patches of size $300 \times 300$ in order to fit in memory and allow batch training, and a margin is added to the crops to avoid border effects.
The training takes advantage of on-the-fly data augmentation strategies, such as rotation ($r \in [-0.2, 0.2] \,  rad$), scaling ($s \in [0.8, 1.2]$) and mirroring, since data augmentation has shown to result in less original data needs.

In practice, setting up the training process is rather easy. 
The training parameters and choices are applicable to most experiments and the only parameter that needs to be chosen is the resizing size of the input image. Indeed, the resolution of the input image needs to be carefully set so that the receptive field of the network is sufficiently large according to the type of task.

All trainings and inferences run on a Nvidia Titan X Pascal GPU.  
Thanks to the pretrained weights used in the contracting path of the network, the training time is significantly reduced.
During the experiments, we also noted that the pretrained weights seemed to help regularization since the model appeared to be less sensitive to outliers.

\section{Experiments}
In order to investigate the performance of the proposed method and to demonstrate its generality, dhSegment is applied on five different tasks related to document processing. Three tasks consisting in page extraction, baseline detection and document segmentation are evaluated and the results are compared against state-of-the art methods. Two additional private datasets are used to show the possibilities and performance of our system on ornaments and photograph extraction. For each task, the reported results are averaged over 5 runs and the best set of post-processing parameters are selected according to the performance on the respective evaluation set. 

\subsection{Page extraction}
\label{sec:page}
Images of digitized historical documents very often include a surrounding border region, which can alter the outputs of document processing algorithms and lead to undesirable results. It is therefore useful to be able to extract the page document from the digitised image. We use the dataset proposed by \cite{tensmeyer2017pagenet} to apply our method and compare our results to theirs in Table \ref{tab:page}. Our method achieves very similar results to human agreement. 

The network is trained to predict for each pixel if it belongs to the main page, essentially predicting a binary mask of the desired page. Training is done on 1635 images for 30 epochs, using a batch size of 1. Since the network should see the image entirely in order to detect the page, full images are used (no patches) but are resized to $6\cdot 10^{5}$ pixels and their aspect ratio is kept. The training takes around 4 hours.

To obtain a binary image from the probabilities output by the network, Otsu's thresholding is applied. Then morphological opening and closing operators are used to clean the binary image. Finally, the quadrilaterals containing the page are extracted by finding the four most extreme corner points of the binary image.

%
\begin{table}[htbp]
\setlength\tabcolsep{4pt}
\caption{Results for the page extraction task (mIoU)}
\begin{center}
\begin{tabular}{|l|c|c|c|}
\hline
Method & cBAD-Train & cBAD-Val & cBAD-Test \\
\hline
Human Agreement & - & 0.978 & 0.983 \\
\hline
Full Image 		& 0.823 & 0.831 & 0.839 \\
Mean Quad 		& 0.833 & 0.891 & 0.894 \\
GrabCut 	\cite{rother2004grabcut}		& 0.900 & 0.906 & 0.916 \\
PageNet \cite{tensmeyer2017pagenet} & 0.971 & 0.968 & 0.974 \\
\bf{dhSegment (quads)} 		& \bf{0.98$\pm6e^{-4}$} & \bf{0.98$\pm8e^{-4}$} & \bf{0.98$\pm8e^{-4}$} \\
\hline 
\end{tabular}
\label{tab:page}
\end{center}
\vspace*{-\baselineskip}
\end{table}
%

\begin{figure}[ht]
\centering
	\subfloat{\includegraphics[width=.3\linewidth]{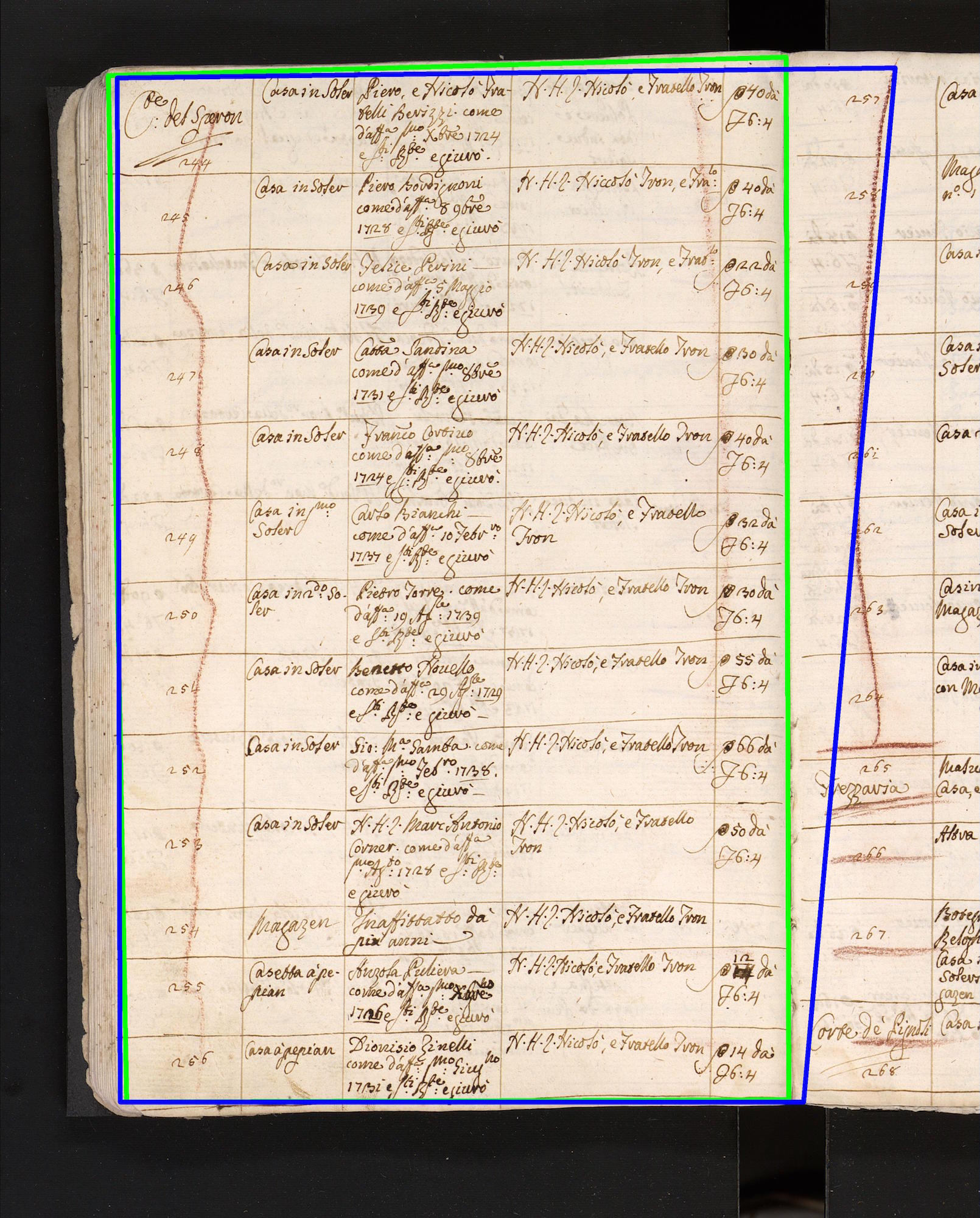}}%
	\hfill
	\subfloat{\includegraphics[width=.3\linewidth]{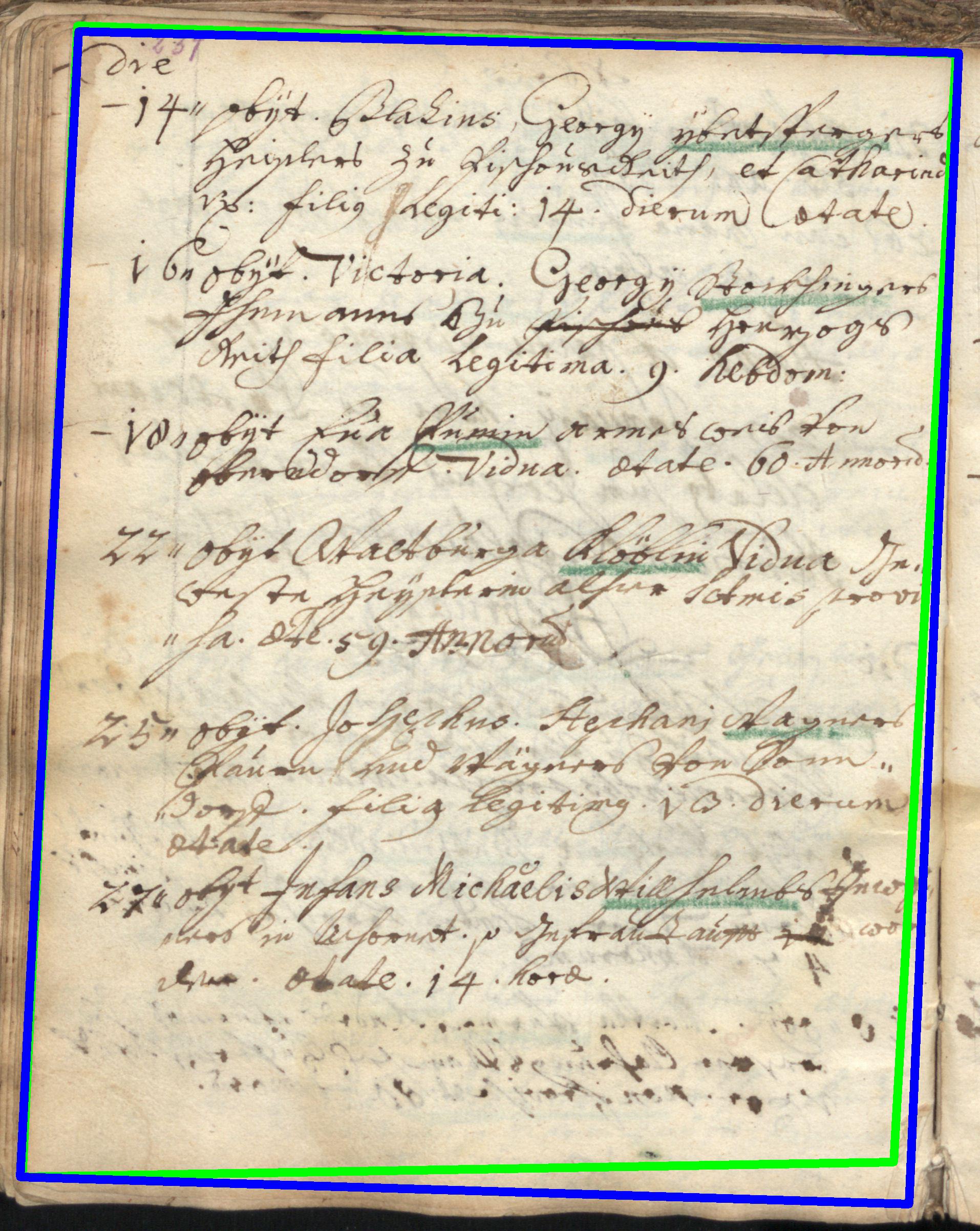}}%
	\hfill
	\subfloat{\includegraphics[width=.3\linewidth]{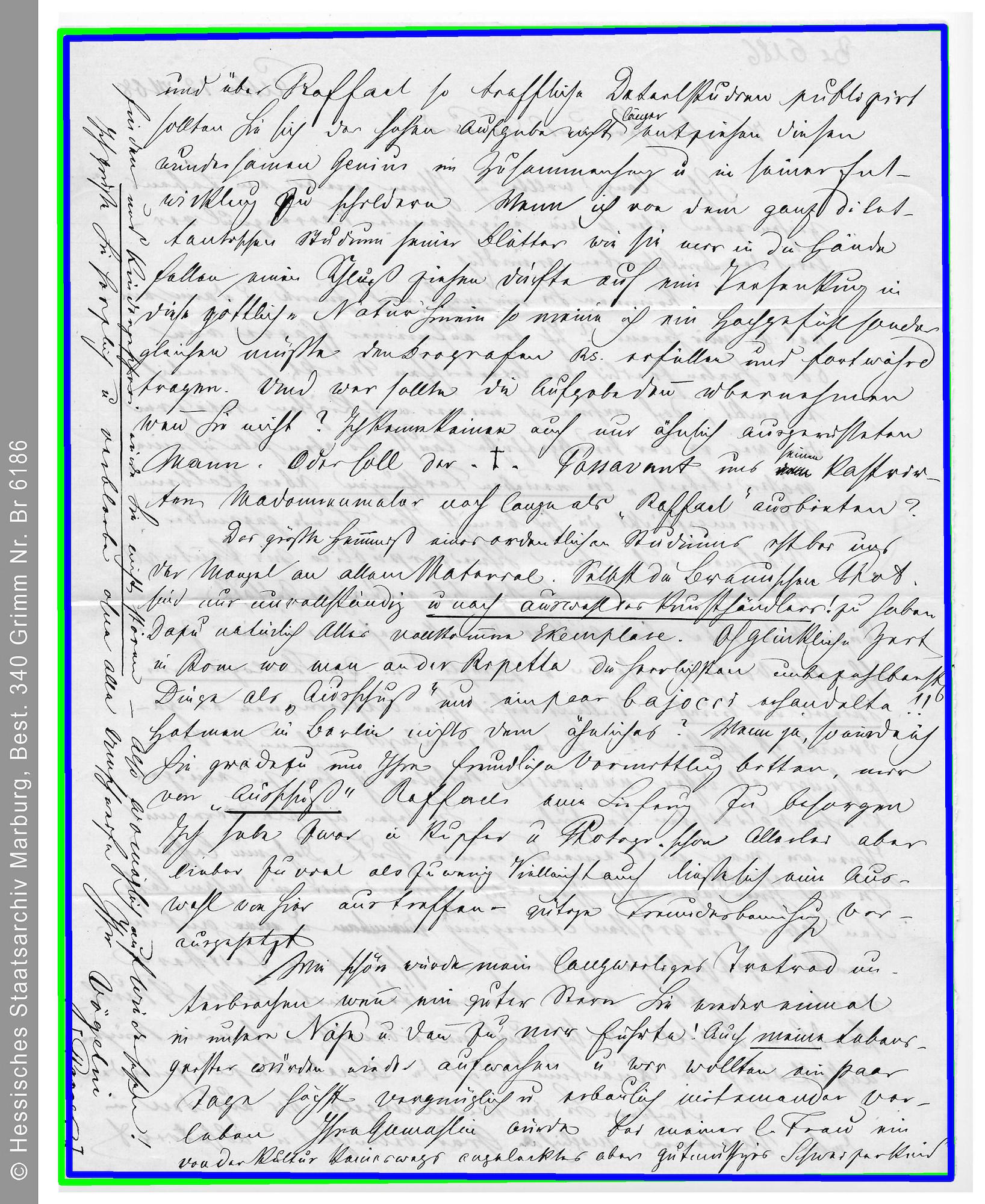}}%
	
\caption{Example of page detection on the cBAD-test set. Green rectangles indicate the ground-truth pages and blue rectangles correspond to the detections generated by dhSegment. The first extraction is inaccurate because part of the side page is also detected. The second one is slightly inaccurate according to the groundtruth but the entire page is still detected. The last example shows a correct detection.}
\label{fig:page}
\end{figure}

\subsection{Baseline detection}
Text line detection is a key step for text recognition applications and thus of great utility in historical document processing. A baseline is defined as a ``virtual line where most characters rest upon and descenders extend below''. The READ-BAD dataset \cite{gruning2017read} has been used for the cBAD: ICDAR2017 Competition \cite{diem2017cbad}.

Here the network is trained to predict the binary mask of 
pixels which are in a small 5-pixel radius of the training baselines.
Each image is resized to have $10^6$ pixels, training is done for 30 epochs, taking around 50 minutes for the complex track of \cite{gruning2017read}.

The probability map is then filtered with a Gaussian filter ($\sigma = 1.5$) before using hysteresis thresholding\footnote{Applying thresholding with $p_{low}$ then only keeping connected components which contains at least a pixel value $p \geq p_{high}$} ($p_{high}=0.4$, $p_{low}=0.2$). The obtained binary mask is decomposed in connected components, and each component is finally converted to a polygonal line.

\begin{table}[htbp]
\setlength\tabcolsep{2pt}
\caption{Results for the cBAD : ICDAR2017 Competition on baseline detection \cite{diem2017cbad} (test set)}
\begin{center}
\vspace*{-\baselineskip}
\hspace*{-14pt}\begin{tabular}{|l|c|c|c|c|c|c|}
\hline
Method & \multicolumn{3}{|c|}{Simple Track} & \multicolumn{3}{|c|}{Complex Track}\\
  		& P-val & R-val & F-val & P-val & R-val & F-val\\
\hline
LITIS & 0.780 & 0.836 & 0.807 & - & - & - \\
IRISA & 0.883 & 0.877 & 0.880 & 0.692 & 0.772 & 0.730 \\
UPVLC & 0.937 & 0.855 & 0.894 & 0.833 & 0.606 & 0.702 \\
BYU & 0.878 & 0.907 & 0.892 & 0.773 & 0.820 & 0.796 \\
DMRZ & \bf{0.973} & \bf{0.970} & \bf{0.971} & \bf{0.854} & 0.863 & \bf{0.859} \\
\bf{dhSeg.} & 0.88$\pm.023$ & \bf{0.97}$\pm.003$ & 0.92$\pm.011$ & 0.79$\pm.021$ & \bf{0.95$\pm.005$} & \bf{0.86$\pm.011$} \\
\hline
\end{tabular}
\label{tab:baseline_cbad}
\end{center}
\vspace*{-\baselineskip}
\end{table}

\begin{figure}[htbp]
	\centering
	\subfloat{\includegraphics[width=.25\linewidth]{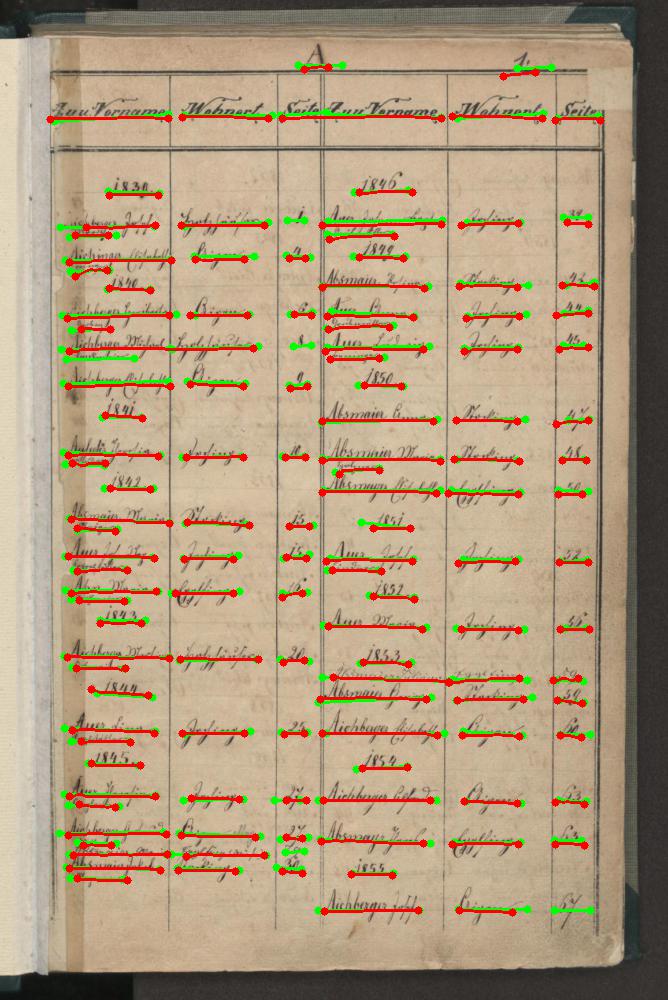}}
	\vspace{0.07cm}
	\subfloat{\includegraphics[width=.25\linewidth]{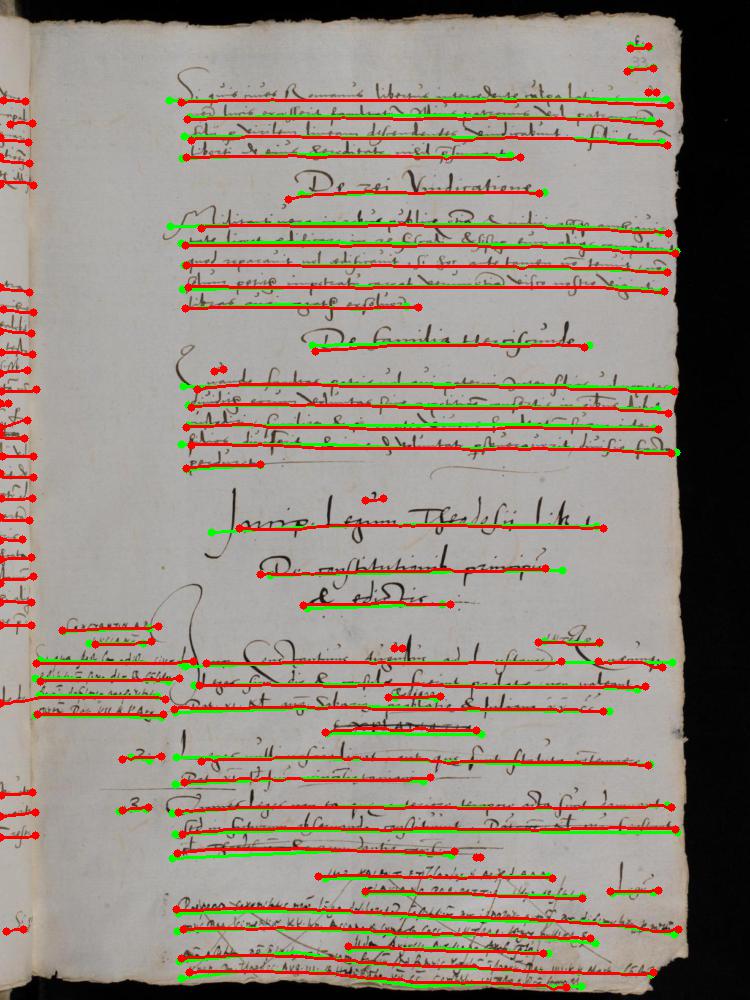}}
	\vspace{0.07cm}
	\subfloat{\includegraphics[width=.47\linewidth]{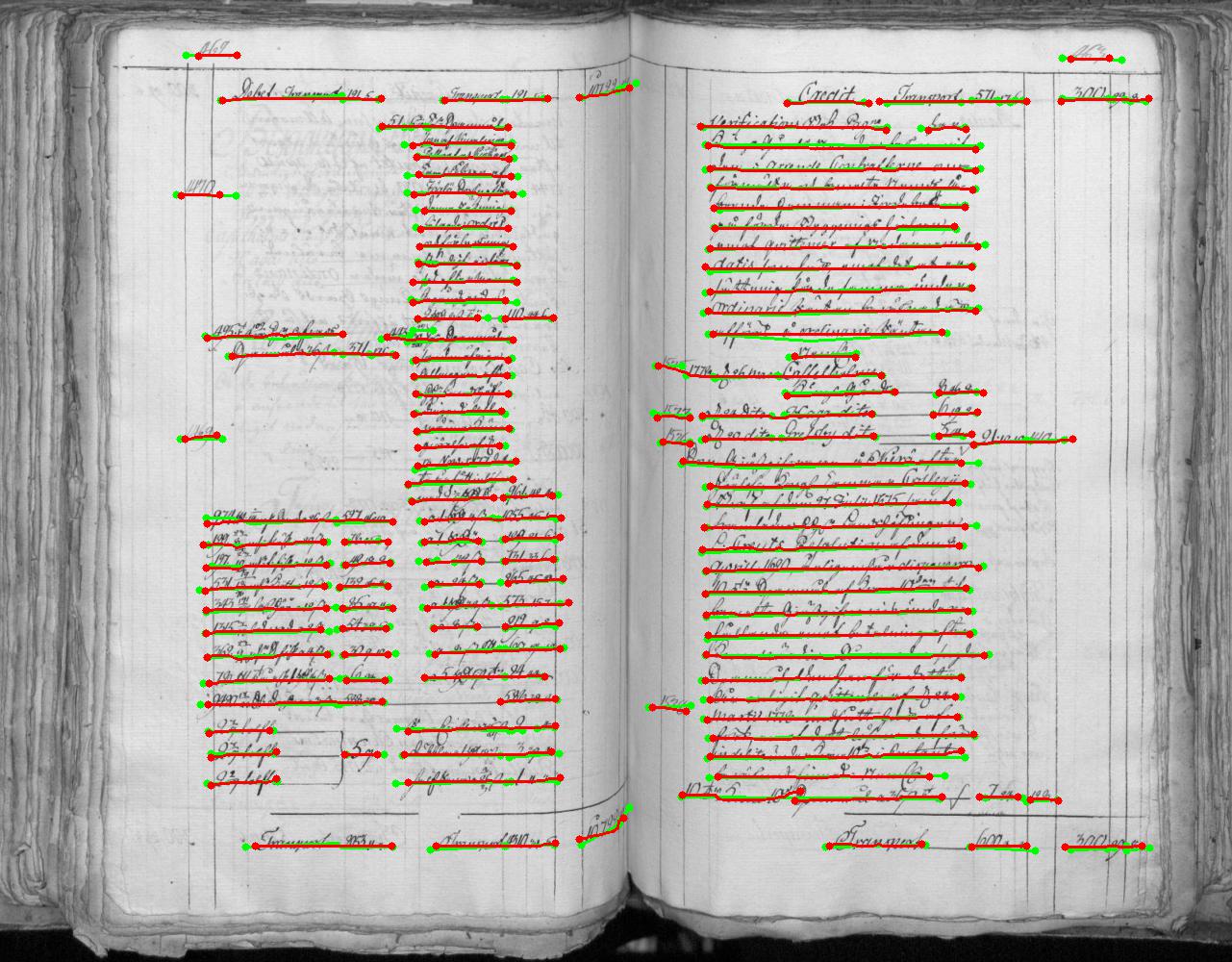}}

\caption{Examples of baseline extraction on the complex track of the cBAD dataset. The ground-truth and predicted baselines are displayed in green and red respectively. Some limitations of the simple approach we propose can be seen here, for instance detecting text on the neighbouring page (right), or merging close text lines together (left and middle). These issues could be addressed with a more complex pipeline incorporating, for instance, page segmentation (as seen in Section \ref{sec:page}) or by having the network predict additional features, but this goes beyond the scope of this paper.}
\label{fig:baseline}
\end{figure}

\subsection{Document layout analysis}

Document Layout Analysis refers to the task of segmenting a given document into semantically meaningful regions. In the experiment, we use the DIVA-HisDB dataset \cite{simistira2016diva} and perform the task formulated in \cite{simistira2017icdar2017}. The dataset is composed of three manuscripts with 20 training, 10 evaluation and 10 testing images for each manuscript. In this task, the layout analysis focuses on assigning each pixel a label among the following classes : text regions, decorations, comments and background, with the possibility of multi-class labels (e.g a pixel can be part of the main-text-body but at the same time be part of a decoration). Our results are compared with the participants of the competition in Table \ref{tab:diva}.

For each manuscript, a model is trained solely on the corresponding 20 training images for 30 epochs. No resizing of the input images is performed but patch cropping is used to allow for batch training. A batch size of $8$ and patches of size $400 \times 400$ are used for manuscripts CSG18 and CSG863 but because the images of manuscript CB55 have higher resolution (approximately a factor 1.5) the patch size is increased to $600 \times 600$ and the batch size is reduced to $4$ to fit into memory. The training of each model lasts between two and four hours.

The post-processing consists in obtaining a binary mask for each class and removing small connected components. The threshold is set to $0.5$ and the components smaller than $50$ pixels are removed. The mask obtained by the page detection (Section \ref{sec:page}) is also used as post-processing to improve the results, especially to reduce the false positive text detections on the borders of the image.

\begin{table}[htbp]
\caption{Results for the ICDAR2017 Competition on Layout Analysis for Challenging Medieval Manuscripts \cite{simistira2017icdar2017} - Task-1 (IoU)}
\begin{center}
\setlength\tabcolsep{3pt}
\vspace*{-\baselineskip}
\begin{tabular}{|l|c|c|c|c|}
\hline
Method & CB55 & CSG18 & CSG863 & Overall \\
\hline
System-1 (KFUPM) & .7150 & .6469 & .5988 & .6535 \\
System-6 (IAIS) & .7178 & .7496 & .7546 & .7407 \\
System-4.2 (MindGarage-2) & .9366 & .8837 & .8670 & .8958 \\
System-2 (BYU) & .9639 & .8772 & .8642 & .9018 \\
System-3 (Demokritos) & .9675 & .9069 & .8936 & .9227 \\
\bf{dhSegment} 	& .974$\pm.001$ & .928$\pm.002$ & .905$\pm.007$ & .936$\pm.004$ \\ 
\bf{dhSegment + Page} 	& .978$\pm.001$  & .929$\pm.002$ & .909$\pm.006$ & .938$\pm.004$ \\ 
System-4.1 (MindGarage-1) & .9864 & .9357 & .8963 & .9395 \\
System-5 (NLPR) & .9835 & .9365 & .9271 & .9490 \\
\hline 
\end{tabular}
\label{tab:diva}
\end{center}
\vspace*{-\baselineskip}
\end{table}

\begin{figure}[htbp]
\centering
	\subfloat{\includegraphics[trim=0 900 450 150,clip, width=.3\linewidth]{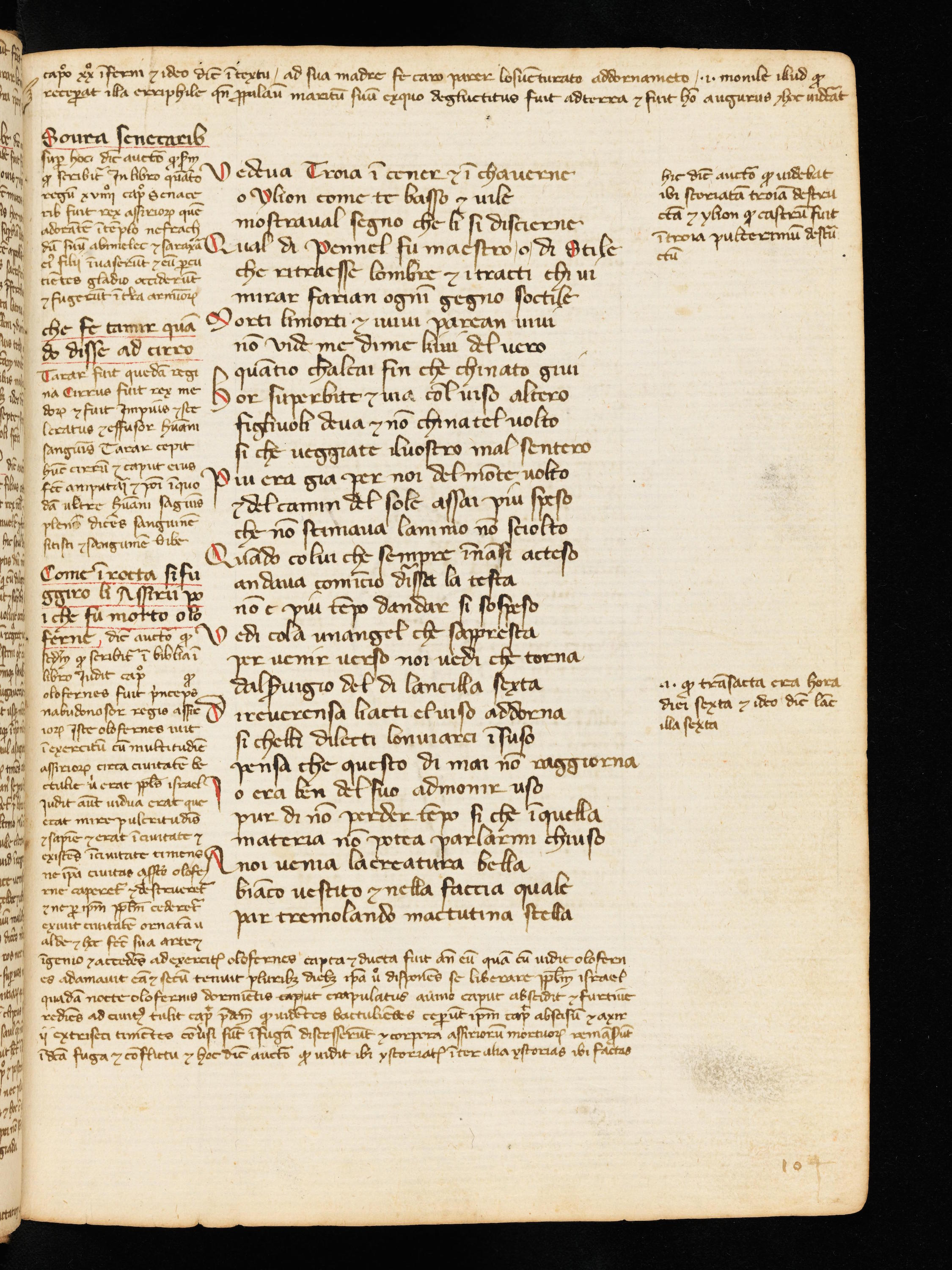}}
	\hfill
	\subfloat{\includegraphics[trim=0 900 450 150,clip, width=.3\linewidth]{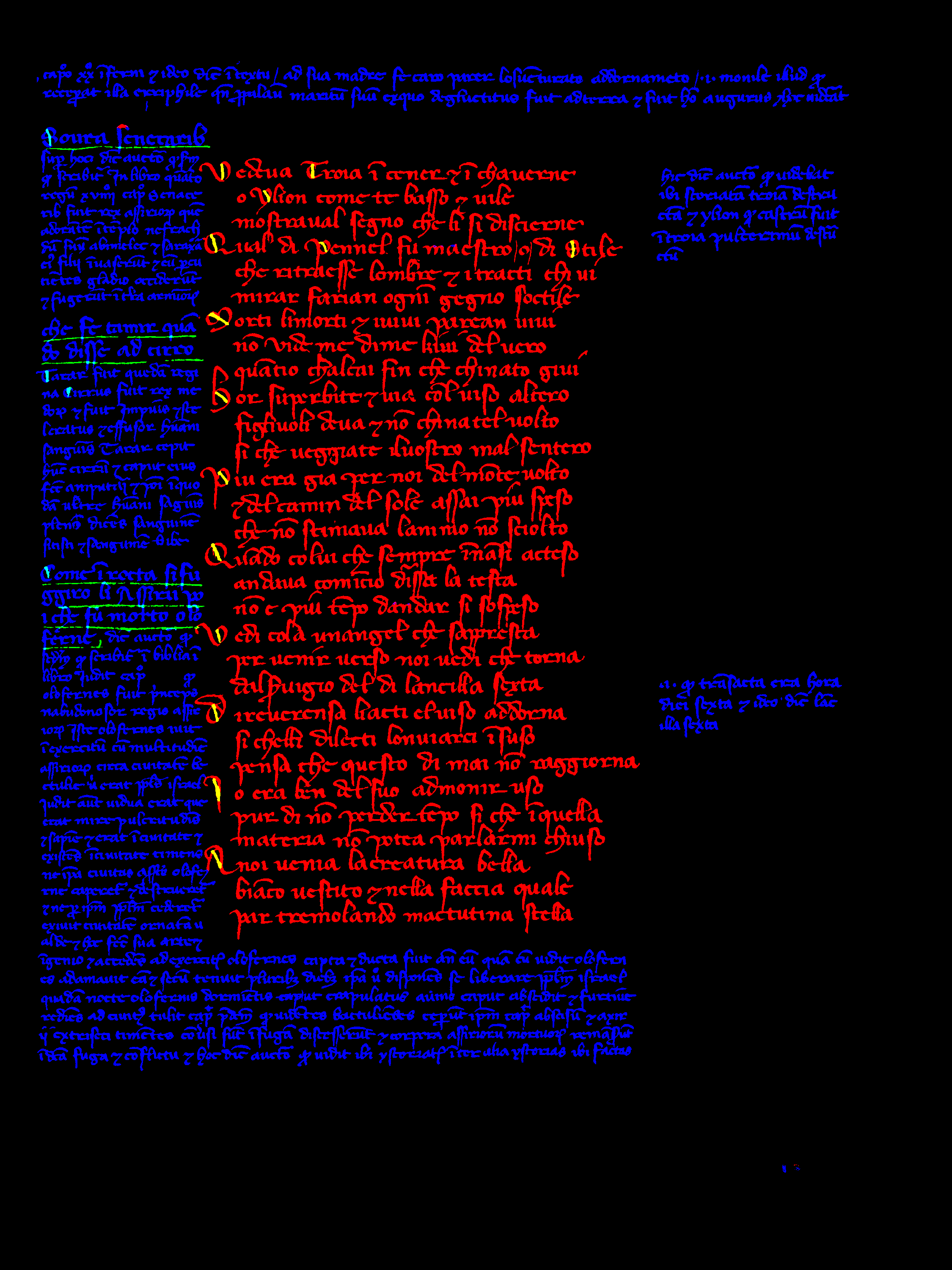}}
	\hfill
	\subfloat{\includegraphics[trim=0 900 450 150,clip, width=.3\linewidth]{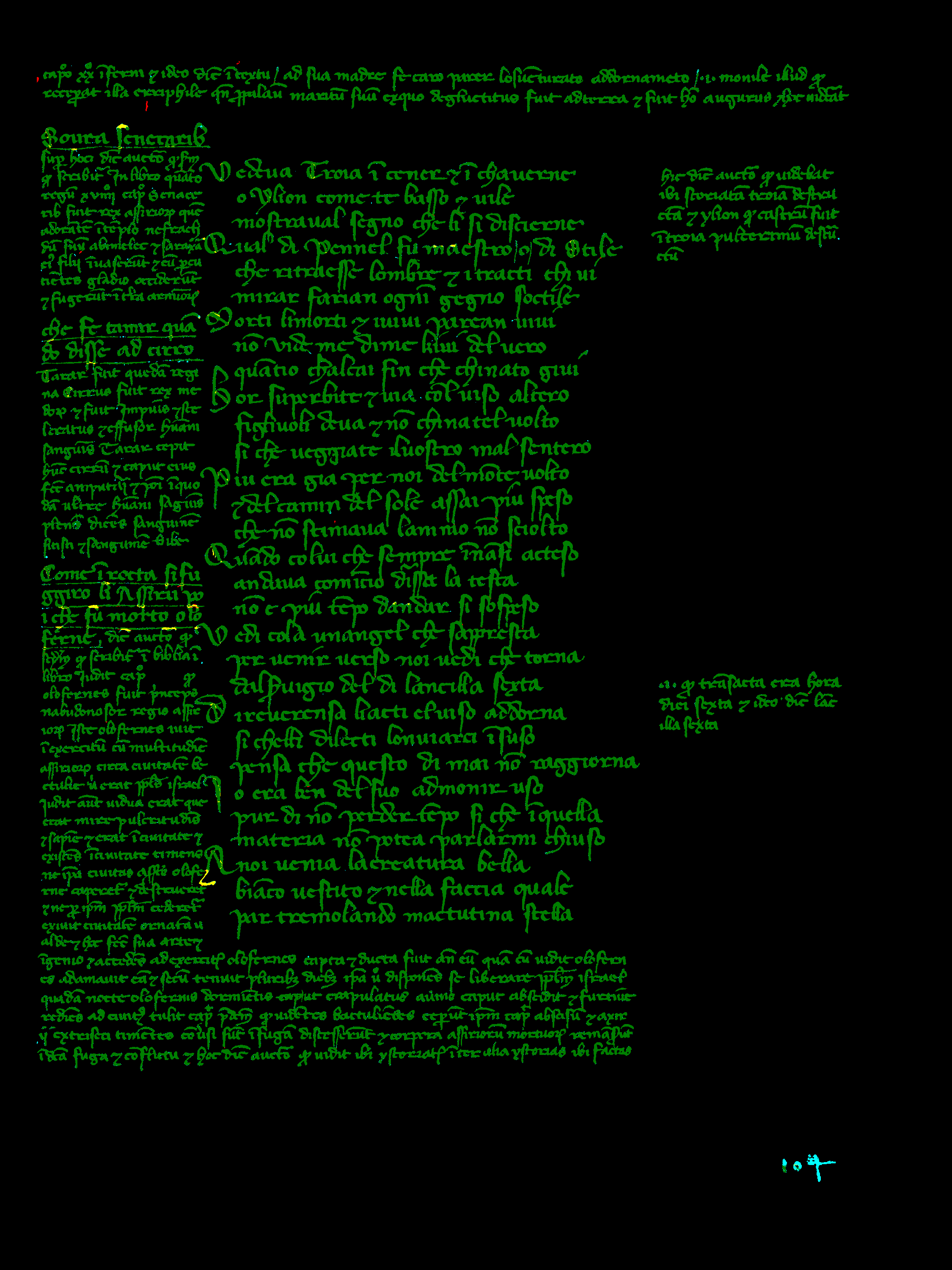}}

\caption[caption footnote]{Example of layout analysis on the DIVA-HisDB test test. On the left the original manuscript image, in the middle the classes pixel-wise labelled by the dhSegment and on the right the comparison with the ground-truth (refer to the evaluation tool\protect\footnotemark for the signification of colors)}
\label{fig:diva}
\end{figure}
\footnotetext{https://github.com/DIVA-DIA/DIVA\_Layout\_Analysis\_Evaluator}

\subsection{Ornament detection}
Ornaments are decorations or embellishments which can be found in many manuscripts. The study of ornaments and discovery of unexpected details is often of major interest for historians. Therefore a system capable of filtering the pages containing such decorations in large collections and locate their positions is of great assistance. 

The private dataset\cite{JunkerThesis} used for this task is composed of several printed books. The selected pages were manually annotated and each ornament was marked by a bounding rectangle. A total of 912 annotated pages were produced, with 612 containing one or several ornaments. The dataset is split in the following way :
\begin{itemize}
	\item 610 pages for training (427 with ornaments)
	\item 92 pages for evaluation (62 with ornaments)
	\item 183 pages for testing (123 with ornaments)
\end{itemize}

The original images are resized to $8 \cdot 10^5$ and the model is trained for 30 epochs with batch size of $16$. The training takes less than two hours. 

To obtain the binary masks, a threshold of $t=0.6$ is applied to the output of the network. Opening and closing operations are performed and a bounding rectangle is fitted to each detected ornament. Finally, very small boxes (those with areas less than 0.5\% of the image size) are removed. 

The evaluation of ornaments detection task is measured using the standard Intersection over Union (IoU) metric, which calculates how well the predicted and the correct boxes overlap. Table \ref{tab:ornaments} lists the precision, recall and f-measure for three IoU thresholds as well as the mean IoU (mIoU) measure. Our results are compared to the method implemented in \cite{JunkerThesis} which uses a region proposal technique coupled with a CNN classifier to filter false positives.

\begin{figure}[htbp]
\centering
	\subfloat{\includegraphics[trim=50 50 50 50,clip, width=.3\linewidth]{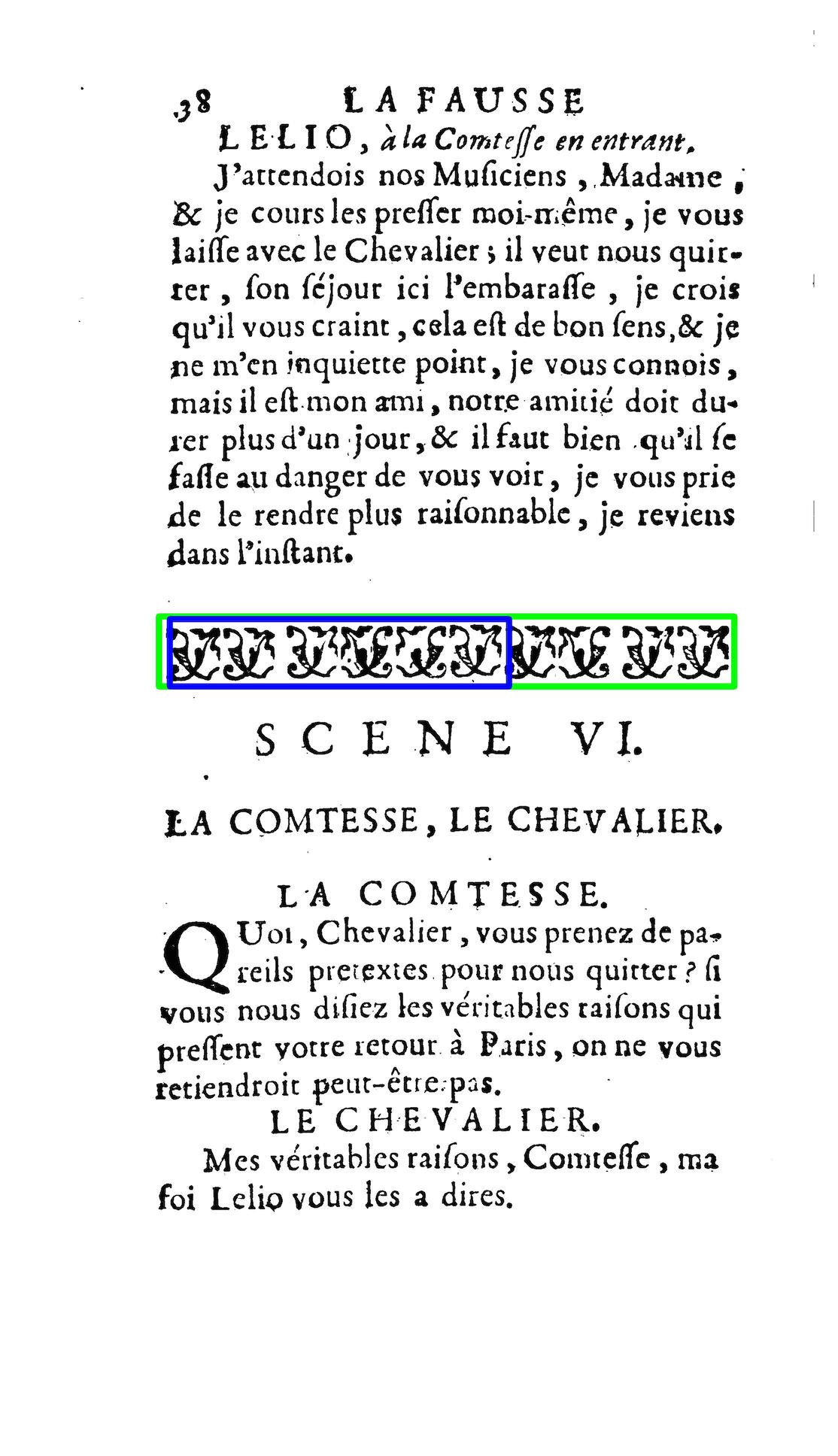}}
	\hfill
	\subfloat{\includegraphics[trim=250 250 250 250,clip, width=.3\linewidth]{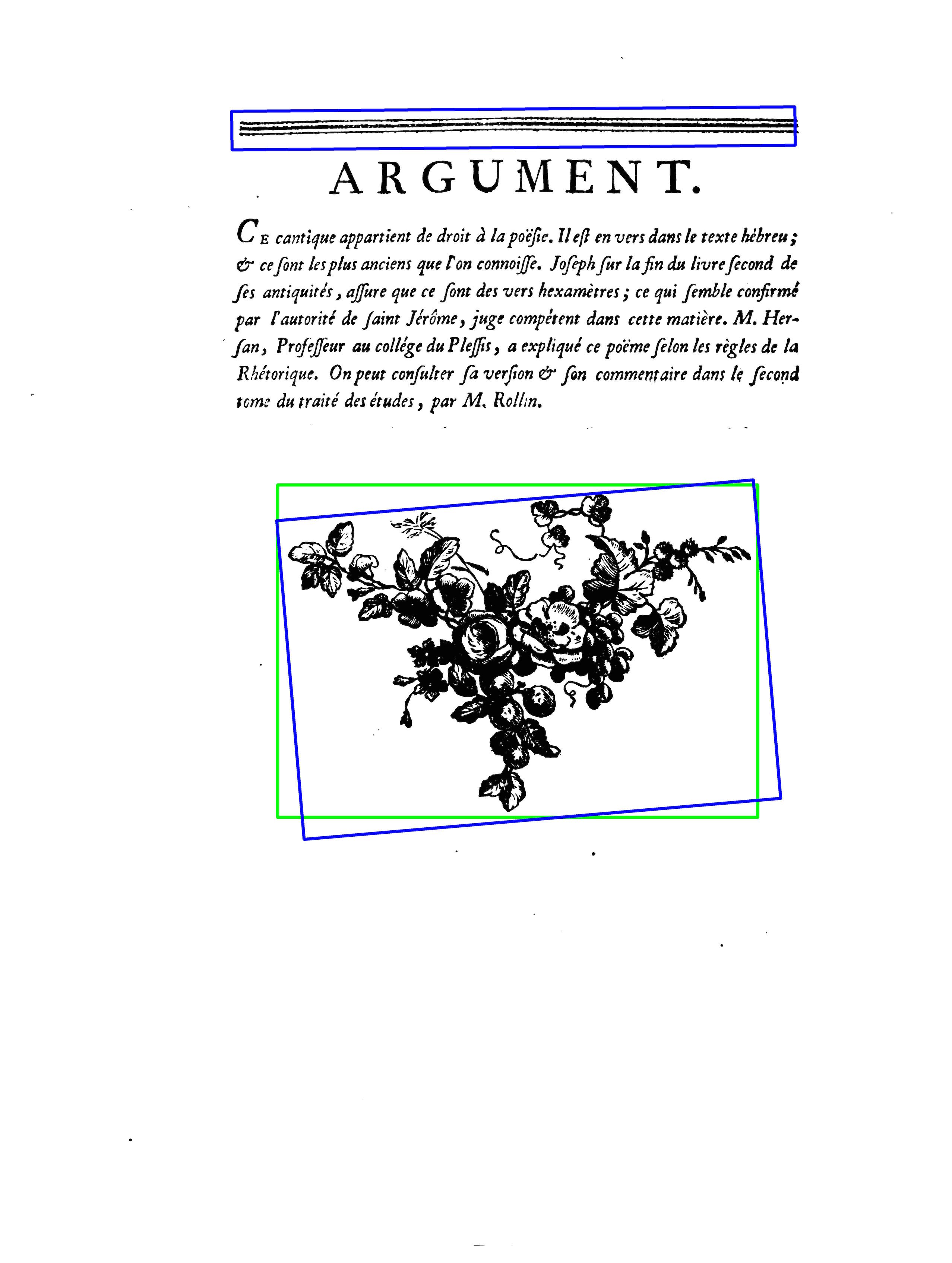}}
	\hfill
	\subfloat{\includegraphics[trim=100 100 100 100,clip, width=.3\linewidth]{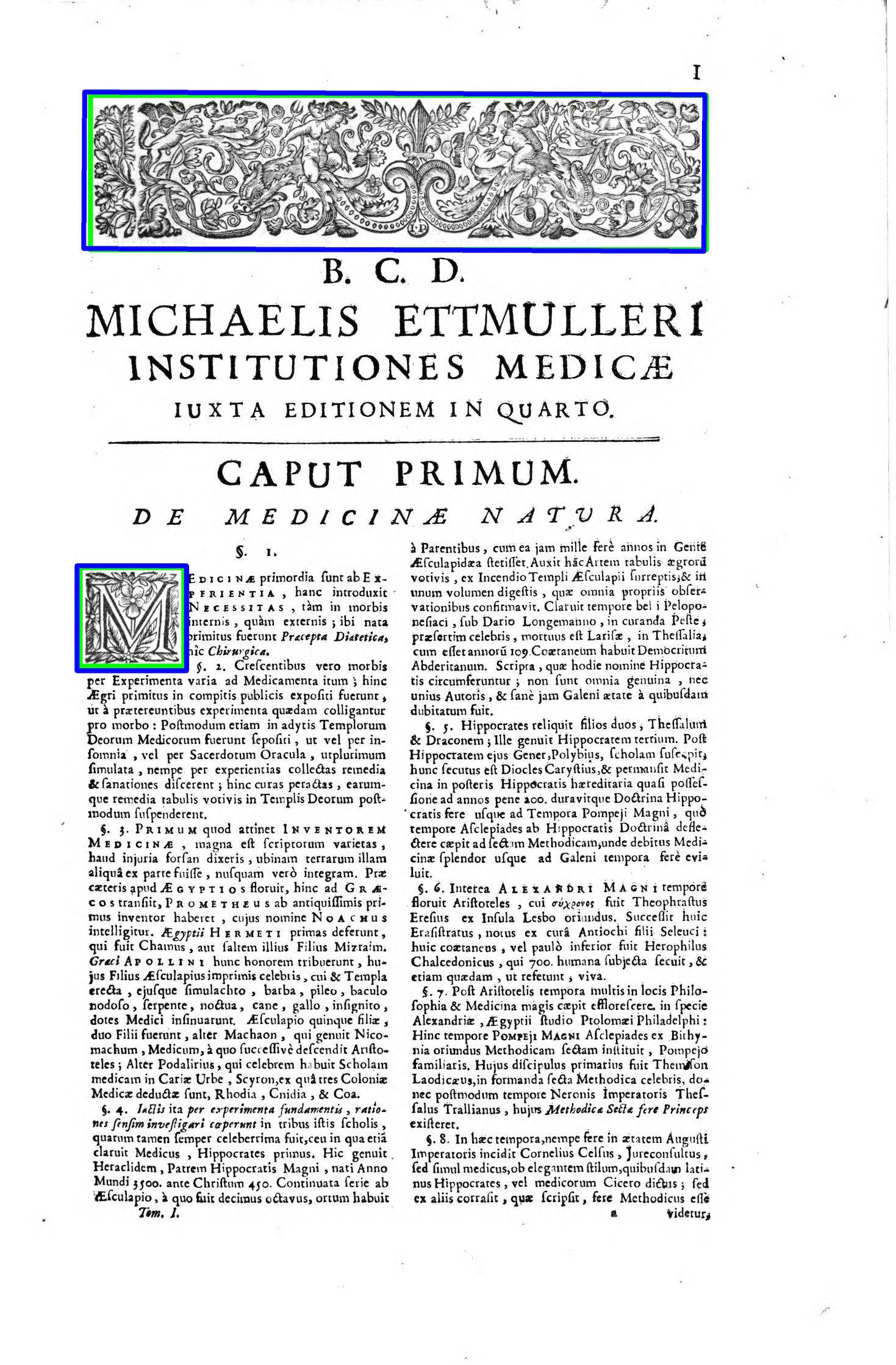}}
	
\caption{The left image illustrates the case of a partially detected ornament, the middle one shows the detection of an illustration but also a false positive detection of the banner and the right image is a correct example of multiple ornaments extraction.}
\label{fig:ornaments}
\end{figure}

%
%
\begin{table}[htbp]
\setlength\tabcolsep{4pt}
\caption{Ornaments detection task. Evaluation at different IoU thresholds on test set}
\begin{center}
\begin{tabular}{| l | c | c c c | c |}

\hline
Method & IoU & F-val & P-val & R-val & mIoU\\
\hline
\cite{JunkerThesis}-config1 & 0.5 & 0.560 & 0.800 & 0.430 & - \\
\cite{JunkerThesis}-config2 & 0.5 & 0.527 & 0.470 & 0.600 & - \\
\hline
\multirow{3}{*}{dhSegment} & 0.7 & 0.94$\pm$.023 & 0.96$\pm$.036 & 0.92$\pm$.013 & \\ 
 & 0.8 & 0.87$\pm$.033 & 0.84$\pm$.049 & 0.91$\pm$.016 & 0.87$\pm$.016 \\
 & 0.9 & 0.56$\pm$.054 & 0.42$\pm$.053 & 0.83$\pm$.036 & \\
\hline
\end{tabular}
\label{tab:ornaments}
\end{center}
\vspace*{-\baselineskip}
\end{table}

\subsection{Photo-collection extraction}

A very practical case comes from the processing of the scans of an old photo-collection. The inputs are high resolution scans of pieces of cardboard with an old photograph stuck in the middle, and the task is to properly extract the part of the scan containing the cardboard and the image respectively.

Annotation was done very quickly by directly drawing on the scans the part to be extracted in different colors (background, cardboard, photograph). Leveraging standard image editing software, 60 scans per hour can be annotated. Data was split in 100 scans for training, 20 for validation and 150 for testing. Training for 40 epochs took only 20 minutes. The network is trained to predict for each pixel its belonging to one of the classes.

The predicted classes are then cleaned with a simple morphological opening, and the smallest enclosing rectangle of the corresponding region is extracted. Additionally, one can use the layout constraint that the area of the photograph has to be enclosed in the area of the piece of cardboard. We compare the extracted rectangle with the smallest rectangle coming from the groundtruth mask and display relevant metrics in Table~\ref{tab:cini}.

\begin{table}[htbp]
\caption{Photo-collection extraction task. Evaluation of mIoU on test set, and some Recall at IoU thresholds of 0.85 and 0.95}
\begin{center}
\begin{tabular}{|l|c |c c c|}

\hline
Method & Cardboard & \multicolumn{3}{|c|}{Photo} \\
 & mIoU & mIoU & R@0.85 & R@0.95 \\
\hline
Predictions-only  & 0.992 & 0.982 & 0.980 & 0.967\\
+ layout constraint & 0.992 & 0.988 & 1.000 & 0.993\\
\hline
\end{tabular}
\label{tab:cini}
\end{center}
\vspace*{-\baselineskip}
\end{table}

\begin{figure}[htbp]
\centering
	\subfloat{\includegraphics[width=.43\linewidth]{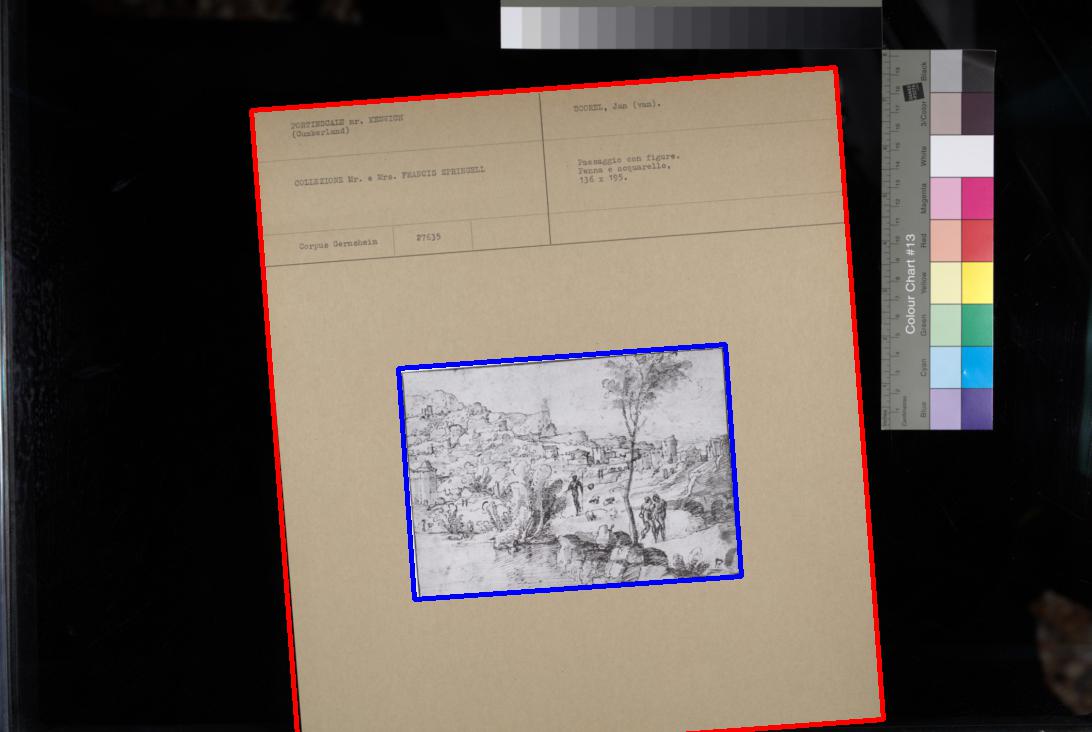}}
	\hfill
	\subfloat{\includegraphics[width=.43\linewidth]{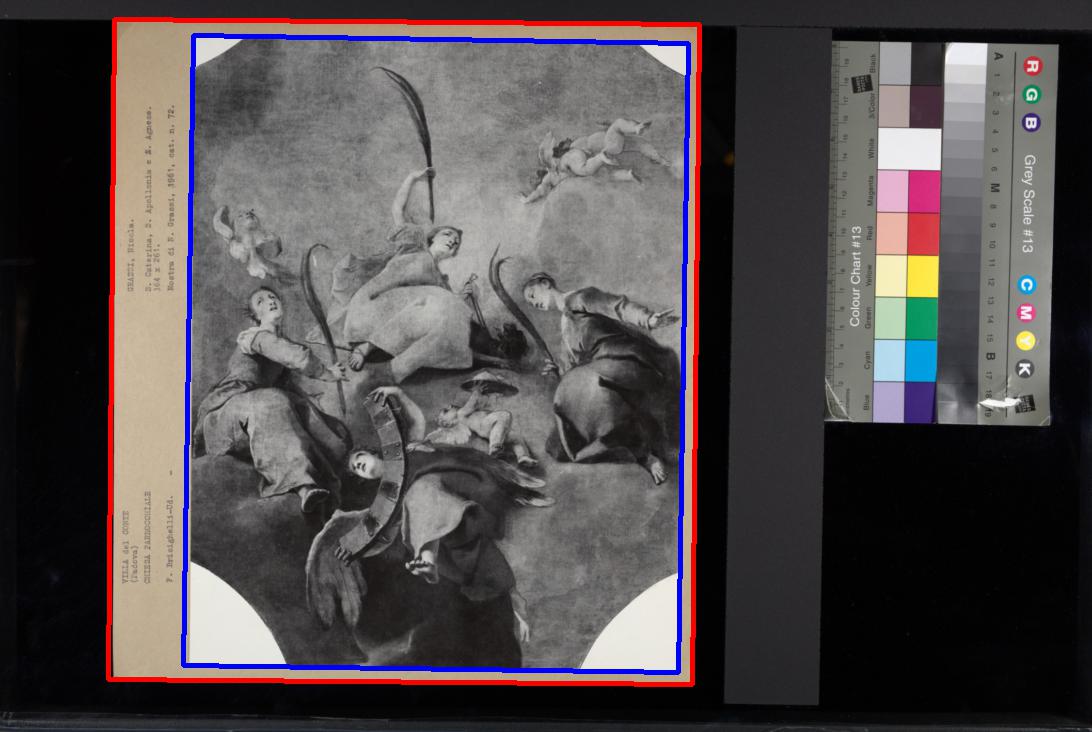}}
\caption{Example of extraction on the photo-collection scans. Note that contrary to the ornaments case, the zone to be extracted is better defined which allow for a much more precise extraction.}
\label{fig:cini}
\end{figure}

\section{Discussion}

While using the same network and almost the same training configurations, the results we obtained on the five different tasks evaluated are competitive or better than the state-of-the-art. Despite the genericity and flexibility of the approach, we can also highlight the speed of the training (less than an hour in some cases), as well as the little amount of training data needed, both thanks to the pre-trained part of the network.

The results presented in this paper demonstrate that a generic deep learning architecture, retrained for specific segmentation tasks using a standardized process, can, in certain cases, outperform dedicated systems. This has multiple consequences.

First, it opens an avenue towards simple, off-the-shelf, programming bricks that can be trained by non-specialists and used for large series of document analysis problems. Using such generic bricks, one just has to design examples of target masks to train a dedicated segmentation system. Such bricks could easily be integrated in intuitive visual programming environments to take part in more complex pipelines of document analysis processes.


Eventually, although the scenario discussed in this article presents how the \emph{same} architecture can be trained to perform efficiently on separate segmentation tasks, further research should study how incremental or parallel training over various tasks could boost transfer learning for segmentation problems. In other words, it is not unlikely that even better performances could be reached if the same network would try to learn various segmentation tasks simultaneously, instead of being trained in only one kind of problem. From this perspective, the results presented in this paper constitute a first step towards the development of a highly efficient universal segmentation engine.

\section*{Acknowledgment}
This work was partially funded by the European Union's Horizon 2020 research and innovation programme under grant agreement No 674943 (READ Recognition and Enrichment of Archival Documents)

\bibliographystyle{ieeetr}
\bibliography{content/bib}

\begin{thebibliography}{10}

\bibitem{long2015fully}
J.~Long, E.~Shelhamer, and T.~Darrell, ``Fully convolutional networks for
  semantic segmentation,'' in {\em Proceedings of the IEEE conference on
  computer vision and pattern recognition}, pp.~3431--3440, 2015.

\bibitem{deng2009imagenet}
J.~Deng, W.~Dong, R.~Socher, L.-J. Li, K.~Li, and L.~Fei-Fei, ``Imagenet: A
  large-scale hierarchical image database,'' in {\em Computer Vision and
  Pattern Recognition, 2009. CVPR 2009. IEEE Conference on}, pp.~248--255,
  IEEE, 2009.

\bibitem{ronneberger2015u}
O.~Ronneberger, P.~Fischer, and T.~Brox, ``U-net: Convolutional networks for
  biomedical image segmentation,'' in {\em International Conference on Medical
  image computing and computer-assisted intervention}, pp.~234--241, Springer,
  2015.

\bibitem{krizhevsky2012imagenet}
A.~Krizhevsky, I.~Sutskever, and G.~E. Hinton, ``Imagenet classification with
  deep convolutional neural networks,'' in {\em Advances in neural information
  processing systems}, pp.~1097--1105, 2012.

\bibitem{simonyan2014very}
K.~Simonyan and A.~Zisserman, ``Very deep convolutional networks for
  large-scale image recognition,'' {\em arXiv preprint arXiv:1409.1556}, 2014.

\bibitem{he2016deep}
K.~He, X.~Zhang, S.~Ren, and J.~Sun, ``Deep residual learning for image
  recognition,'' in {\em Proceedings of the IEEE conference on computer vision
  and pattern recognition}, pp.~770--778, 2016.

\bibitem{simistira2017icdar2017}
F.~Simistira, M.~Bouillon, M.~Seuret, M.~W{\"u}rsch, M.~Alberti, R.~Ingold, and
  M.~Liwicki, ``Icdar2017 competition on layout analysis for challenging
  medieval manuscripts,'' in {\em Document Analysis and Recognition (ICDAR),
  2017 14th IAPR International Conference on}, vol.~1, pp.~1361--1370, IEEE,
  2017.

\bibitem{diem2017cbad}
M.~Diem, F.~Kleber, S.~Fiel, T.~Gruning, and B.~Gatos, ``cbad: Icdar2017
  competition on baseline detection,'' in {\em 2017 14th IAPR International
  Conference on Document Analysis and Recognition (ICDAR)}, vol.~01,
  pp.~1355--1360, Nov. 2017.

\bibitem{pratikakis2017icdar2017}
I.~Pratikakis, K.~Zagoris, G.~Barlas, and B.~Gatos, ``Icdar2017 competition on
  document image binarization (dibco 2017),'' in {\em Document Analysis and
  Recognition (ICDAR), 2017 14th IAPR International Conference on}, vol.~1,
  pp.~1395--1403, IEEE, 2017.

\bibitem{chen2017convolutional}
K.~Chen, M.~Seuret, J.~Hennebert, and R.~Ingold, ``Convolutional neural
  networks for page segmentation of historical document images,'' in {\em
  Document Analysis and Recognition (ICDAR), 2017 14th IAPR International
  Conference on}, vol.~1, pp.~965--970, IEEE, 2017.

\bibitem{xu2017page}
Y.~Xu, W.~He, F.~Yin, and C.-L. Liu, ``Page segmentation for historical
  handwritten documents using fully convolutional networks,'' in {\em Document
  Analysis and Recognition (ICDAR), 2017 14th IAPR International Conference
  on}, vol.~1, pp.~541--546, IEEE, 2017.

\bibitem{breuel2017robust}
T.~M. Breuel, ``Robust, simple page segmentation using hybrid convolutional
  mdlstm networks,'' in {\em Document Analysis and Recognition (ICDAR), 2017
  14th IAPR International Conference on}, vol.~1, pp.~733--740, IEEE, 2017.

\bibitem{tensorflow2015-whitepaper}
M.~Abadi, A.~Agarwal, P.~Barham, E.~Brevdo, Z.~Chen, C.~Citro, G.~S. Corrado,
  A.~Davis, J.~Dean, M.~Devin, S.~Ghemawat, I.~Goodfellow, A.~Harp, G.~Irving,
  M.~Isard, Y.~Jia, R.~Jozefowicz, L.~Kaiser, M.~Kudlur, J.~Levenberg,
  D.~Man\'{e}, R.~Monga, S.~Moore, D.~Murray, C.~Olah, M.~Schuster, J.~Shlens,
  B.~Steiner, I.~Sutskever, K.~Talwar, P.~Tucker, V.~Vanhoucke, V.~Vasudevan,
  F.~Vi\'{e}gas, O.~Vinyals, P.~Warden, M.~Wattenberg, M.~Wicke, Y.~Yu, and
  X.~Zheng, ``{TensorFlow}: Large-scale machine learning on heterogeneous
  systems,'' 2015.
\newblock Software available from tensorflow.org.

\bibitem{nair2010rectified}
V.~Nair and G.~E. Hinton, ``Rectified linear units improve restricted boltzmann
  machines,'' in {\em Proceedings of the 27th international conference on
  machine learning (ICML-10)}, pp.~807--814, 2010.

\bibitem{otsu1979threshold}
N.~Otsu, ``A threshold selection method from gray-level histograms,'' {\em IEEE
  transactions on systems, man, and cybernetics}, vol.~9, no.~1, pp.~62--66,
  1979.

\bibitem{serra1983image}
J.~Serra, {\em Image analysis and mathematical morphology}.
\newblock Academic Press, Inc., 1983.

\bibitem{glorot2010understanding}
X.~Glorot and Y.~Bengio, ``Understanding the difficulty of training deep
  feedforward neural networks,'' in {\em Proceedings of the Thirteenth
  International Conference on Artificial Intelligence and Statistics},
  pp.~249--256, 2010.

\bibitem{kingma2014adam}
D.~P. Kingma and J.~Ba, ``Adam: A method for stochastic optimization,'' {\em
  arXiv preprint arXiv:1412.6980}, 2014.

\bibitem{ioffe2017batch}
S.~Ioffe, ``Batch renormalization: Towards reducing minibatch dependence in
  batch-normalized models,'' in {\em Advances in Neural Information Processing
  Systems}, pp.~1942--1950, 2017.

\bibitem{tensmeyer2017pagenet}
C.~Tensmeyer, B.~Davis, C.~Wigington, I.~Lee, and B.~Barrett, ``Pagenet: Page
  boundary extraction in historical handwritten documents,'' in {\em
  Proceedings of the 4th International Workshop on Historical Document Imaging
  and Processing}, pp.~59--64, ACM, 2017.

\bibitem{rother2004grabcut}
C.~Rother, V.~Kolmogorov, and A.~Blake, ``Grabcut: Interactive foreground
  extraction using iterated graph cuts,'' in {\em ACM transactions on graphics
  (TOG)}, vol.~23, pp.~309--314, ACM, 2004.

\bibitem{gruning2017read}
T.~Gr{\"u}ning, R.~Labahn, M.~Diem, F.~Kleber, and S.~Fiel, ``Read-bad: A new
  dataset and evaluation scheme for baseline detection in archival documents,''
  in {\em 2018 13th IAPR International Workshop on Document Analysis Systems
  (DAS)}, pp.~351--356, IEEE, 2018.

\bibitem{simistira2016diva}
F.~Simistira, M.~Seuret, N.~Eichenberger, A.~Garz, M.~Liwicki, and R.~Ingold,
  ``Diva-hisdb: A precisely annotated large dataset of challenging medieval
  manuscripts,'' in {\em Frontiers in Handwriting Recognition (ICFHR), 2016
  15th International Conference on}, pp.~471--476, IEEE, 2016.

\bibitem{JunkerThesis}
F.~Junker, ``Extraction of ornaments from a large collection of books,''
  Master's thesis, EPFL, 2017.

\end{thebibliography}

\end{document}